%% file: latex/acl_latex.tex
\documentclass[11pt]{article}

\usepackage[preprint]{acl}
\usepackage{float}

\usepackage{times}
\usepackage{latexsym}

\usepackage[T1]{fontenc}

\usepackage[utf8]{inputenc}

\usepackage{microtype}

\usepackage{inconsolata}

\usepackage{graphicx}

\usepackage{hyperref}

\usepackage{placeins}

\usepackage{amsmath}
\usepackage{booktabs}

\usepackage{multirow}

\usepackage{caption}
\usepackage{subcaption}

\usepackage{amssymb}

\usepackage{pifont}
\newcommand{\cmark}{\ding{51}}
\newcommand{\xmark}{\ding{55}}

\title{Who and What?\\Using Linguistic Features and Annotator Characteristics to Analyze Annotation Variation}

\author{
Maximilian Maurer\textsuperscript{1,2}, 
Maximilian Linde\textsuperscript{1} \and 
Gabriella Lapesa\textsuperscript{1,2} \\
\textsuperscript{1}GESIS - Leibniz Institute for the Social Sciences \\ 
\textsuperscript{2}Heinrich-Heine University Düsseldorf\\
\textsuperscript{1}\texttt{first.last@gesis.org}
}

\begin{document}
\maketitle
\begin{abstract}
Human label variation has been established as a central phenomenon in NLP: the perspectives different annotators have on the same item need to be embraced. Data collection practices thus shifted towards increasing the annotator numbers and releasing disaggregated datasets, harmful language being most resourced due to its high subjectivity. 
While this resulted in rich information about \textit{who} annotated (sociodemographics, attitudes, etc.), the \textit{what} (e.g., linguistic properties of items), and their interplay has received little attention.
We present the first large-scale analysis of four reference datasets for 
harmful language detection, bringing together annotator characteristics, linguistic properties of the items, and their interactions in a statistically informed picture. We find that interactions are crucial, revealing intersectional effects ignored in previous work, and that a strong role is played by lexical cues and annotator attitudes. Effect patterns, however, vary considerably across datasets. This 
urges caution about generalization and transferability.\footnote{The code of our analyses is available at \url{https://anonymous.4open.science/r/who_and_what-F1C7}}

\textcolor{red}{Disclaimer: This paper contains examples of vulgar expressions and hateful text items.}
\end{abstract}

\section{Introduction}
\input{latex/sections/introduction}

\section{Related Work}
\input{latex/sections/related_work}

\section{Data}
\input{latex/sections/data}

\section{Methods}
\label{sec:methods}

\noindent
\textbf{Annotator Characteristics.} We use all annotator characteristics in the respective datasets.

\noindent
\textbf{Linguistic Features.}
\input{latex/sections/linguistic_features}

\noindent
\textbf{Preprocessing.}
We filter the datasets to include only items that were annotated by at least three annotators who annotated at least 10 items, respectively.
The linguistic features are token-normalized if they are occurrence-count features (e.g., number of nouns). Then, all linguistic features are standardized to have a mean of $0$ and a standard deviation of $1$.
For the annotator characteristics, we remove all missing and "prefer not to answer" annotators, and all annotator IDs potentially mapping to multiple people (i.e., a single annotator ID for multiple conflicting characteristics). We harmonize \textit{gender} and \textit{education} across datasets, and re-code multiple-choice combinations of \textit{race} to keep the number of categories manageable. Finally, we remove annotators with multiple assignments for \textit{sexuality} and \textit{religion}. 
We provide a more detailed description of the preprocessing choices in Appendix \ref{sec:preprocessing-full}.
Table \ref{tab:data} shows the number of items, annotations, and annotators per item after preprocessing per dataset.

\begin{table}
    \centering
    \resizebox{\columnwidth}{!}{
    \begin{tabular}{lrrrc}
         \toprule
         Dataset & Items & Annotations & Annotators & Ann. per Item\\
         \midrule
         \texttt{MHS} & 3,556 & 17,693 & 1,385 & 4.0$\pm$0.2\\
         \texttt{POPQUORN} & 1,500 & 13,036 & 262 & 8.7$\pm$1.0\\
         \texttt{D3CODE} & 4,402 & 139,379 & 4,309 & 31.7$\pm$16.6\\
         \texttt{CTDP} & 97,489 & 221,087 & 10,958 & 4.7$\pm$0.6\\
         \bottomrule
    \end{tabular}}
    \caption{Dataset size, number of annotations, total number of annotators, and mean number of annotators per item $\pm$ st. deviation per dataset after preprocessing.
    }
    \label{tab:data}
\end{table}

\noindent
\textbf{Linguistic Feature Pre-Selection.}
We use a multi-step semi-automatic feature selection method for the linguistic features to reduce multicollinearity and the number of features while retaining interpretability. This process allows us to remove theoretically equivalent or empirically similar features while keeping an expressive set of features.

Firstly, we pre-select features by calculating pairwise Pearson correlations between all the features and retaining those that correlate lower than a threshold of $r < 0.5$ with all other features.

Secondly, the remaining features that correlate with at least one other feature higher than the threshold are then clustered using single-linkage multilevel clustering
using the remaining features' correlation matrix as a similarity matrix. We inspect the resulting clusters for consistency with theoretical expectations (i.e., that features measuring similar properties end up in the same clusters) and pick one feature per cluster\footnote{In practice, we pick a feature well-used in the literature or the one most intuitively interpretable. We explain this in more detail on the example of \texttt{POPQUORN} in Appendix \ref{sec:clustering}.}.

\noindent
\textbf{Regression Modeling.}
We use Bayesian 
multilevel regression with regularization priors because it allows for selecting features among a large number of inter-correlated features. 
We place Horseshoe priors 
\citep{PiironenVehtari2017, CarvalhoPolsonScott2009} on all fixed effects regression coefficients.
This has the effect that small and uncertain effects are aggressively pushed towards $0$ while larger and more certain effects escape this shrinkage \citep{PiironenVehtari2017, VanerpOberskiMulder2019}.

Annotation behavior,
measured by individual annotators' labeling decisions on the respective ordinal scale,
is modeled as a function of the main effects of the linguistic and annotator features
, the interactions among the annotator features
, and the interactions between linguistic and annotator features
. To incorporate the partially cross-classified data structure, random intercepts for the items and annotators are included.
%
%
We treat the outcome of interest as continuous and model it with a Gaussian likelihood
and an identity link function.


Even though the horseshoe prior
aggressively shrinks small and uncertain effects toward 0, the estimates and their credible intervals are not exactly 0.
Therefore, we consider those effects that have 90\% posterior credible intervals not overlapping with 0 as the \textit{surviving} effects.


\section{Comparing Datasets}
\label{sec:ex1-comparison}
\begin{figure}
    \centering
    \includegraphics[width=\linewidth]{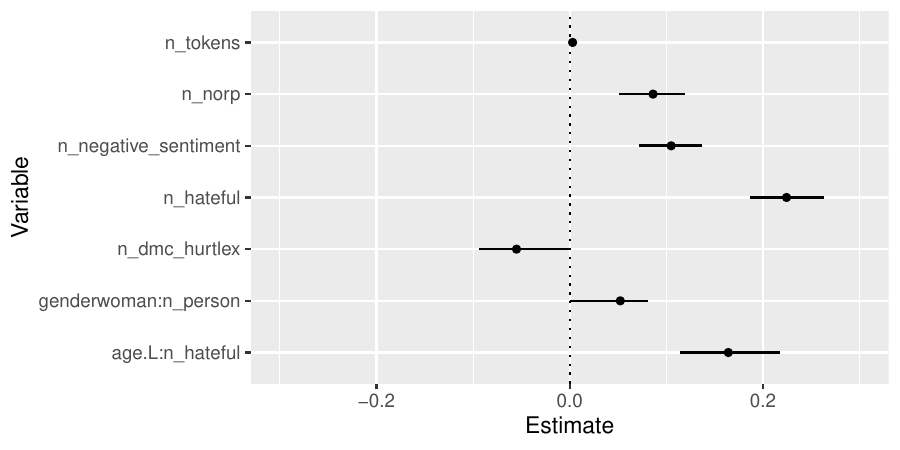}
    \caption{Posterior estimates for the surviving effects for \textsc{POPQUORN}. The dots represent the median posterior estimates, and horizontal bars represent the 95\% highest density interval.}
    \label{fig:popquorn-surv}
\end{figure}

\begin{figure}
    \centering
    \includegraphics[width=\linewidth]{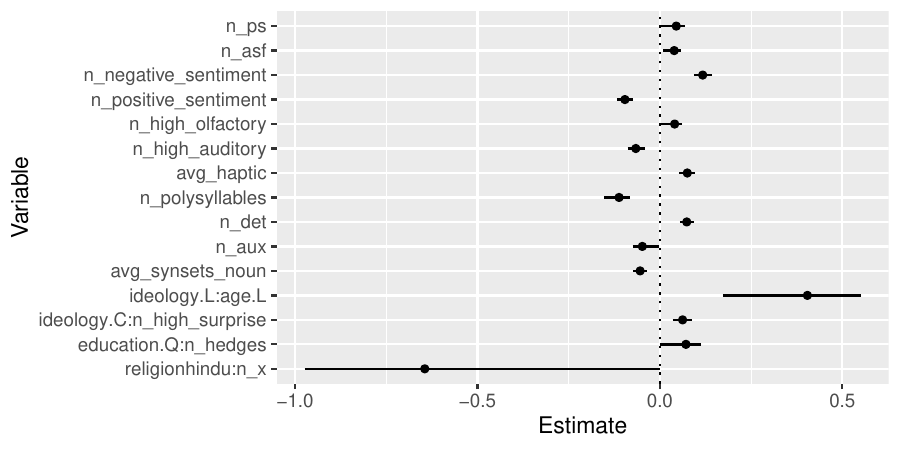}
    \caption{Posterior estimates for the surviving effects for \textsc{MHS}. The dots represent the median posterior estimates, and horizontal bars represent the 95\% highest density interval.}
    \label{fig:mhs-surv}
\end{figure}

For \textsc{MHS} (hate speech) and \textsc{POPQUORN} (offensiveness), we analyze the annotation behavior in the full datasets.
This allows us to compare tendencies in annotation behavior for different datasets for related tasks. Figure \ref{fig:popquorn-surv} shows the surviving effects for \textsc{POPQUORN}, and Figure \ref{fig:mhs-surv} for \textsc{MHS}.

\subsection{POPQUORN}
We find \textbf{no surviving annotator characteristics}\footnote{While this is in contrast to the findings of \citet{pei-jurgens-2023-annotator}, this may be a direct consequence of adding a random intercept for the annotators. For a more in-depth analysis regarding this, see Appendix \ref{sec:reproduction-popqorn}.}.
\textbf{Most surviving effects are linguistic features (5/7)}.
The linguistic survivors we find can be grouped into three components of variation in harmful language annotation: Firstly, explicit and phenomenon-characteristic lexical cues such as negative sentiment (\texttt{n\_negative\_sentiment}) and the presence of hateful, offensive, or vulgar tokens (\texttt{n\_hateful}). 
These results are relevant from a harmful language research perspective, as they confirm assumptions about certain cues and affective dimensions of such language.

Secondly, topical and world knowledge, as indicated by named entities (\texttt{n\_norp}, i.e., nationalities, religious or political groups). Inspection of instances with a high number of such entities reveals that they often mention instances related to controversial topics. We found instances mentioning \textit{Palestinians} and \textit{Israel}) and frequently target ethnic and religious groups (e.g., \textit{Jews},  \textit{Muslims}). On the one hand, such findings are relevant from a human label variation perspective, as they directly tie tendencies in annotations to broader discussions in society. On the other hand, they urge caution from a modeling/content moderation perspective: models may pick up on entity cues, but a functioning content moderation system should not be sensitive to all texts mentioning Muslims or Jews, but rather flag certain topics for human review.  

Thirdly, pragmatic and discourse phenomena such as irony may counterintuitively be related to lexical cues one usually associates with harmful language. For instance, a high number of tokens related to moral or behavioral defects (\texttt{n\_dmc}) is associated with lower offensiveness annotations. Inspection reveals that items with such tokens often are about the author's opposing views on certain positions on moral grounds or are ironic\footnote{See examples in Appendix \ref{sec:text-examples}.}.


We find \textbf{two surviving interactions} between annotator characteristics and linguistic features.
The first indicates differences in age groups for lexical cues (\texttt{age.L:n\_hateful}
, see Figure \ref{fig:pei-interaction-main}\footnote{Find other interaction plots for plots discussed in this paper in Appendix \ref{sec:interactions}.}
)\footnote{For ordinal predictors \texttt{.L} refers to linear, and \texttt{.Q} to quadratic effects.}.
While the presence of hateful words does not correlate with stronger differences in annotation choices at younger ages, the older the annotators get, the more the presence 
of such cues influences annotation choices. Not only does this finding lend itself as an interpretation for potential variation between annotator groups, it also raises questions to the reception of implicit hate speech (i.e., not expressed on the surface) for the different groups.
The second surviving interaction shows gender differences for world knowledge (\texttt{gender:n\_person},
\texttt{n\_person} is the number of proper names such as \textit{Trump}).

\begin{figure}
    \centering
    \includegraphics[width=\linewidth]{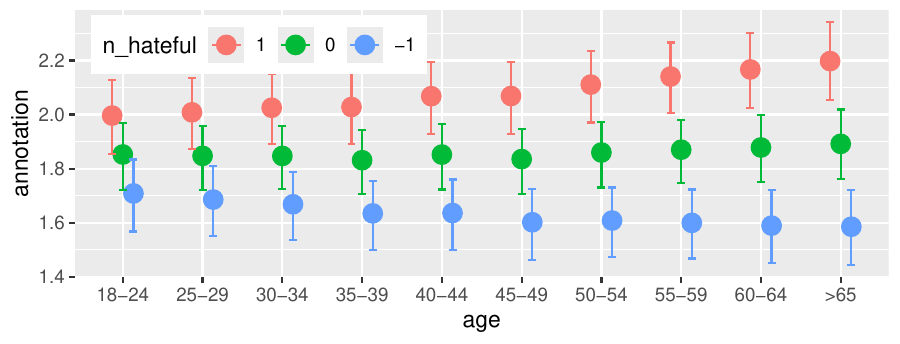}
    \caption{Model predictions for the interaction \texttt{age:n\_hateful} (\textsc{POPQUORN}).
    Labels (\colorbox{red}{1}, \colorbox{green}{0}, \colorbox{cyan}{-1}) refer to SD from mean (0) for \texttt{n\_hateful}. The dots represent the mean posterior estimates, and vertical bars represent the 95\%
    highest density interval.}
    \label{fig:pei-interaction-main}
\end{figure}

\subsection{MHS}

As in \textsc{POPQUORN}, we find \textbf{no surviving annotator characteristics}.
Again, \textbf{most of the surviving effects are linguistic features (11/15)}.
Among them, we find the same high-level patterns as in \textsc{POPQUORN}. Lexical cues show task-specific tendencies: ethnic slurs (\texttt{n\_ps}) and words related to female genitalia (\texttt{n\_asf}). Similarly, inspection of items with more tokens with a high auditory (\texttt{n\_high\_auditory}) and olfactory grounding (\texttt{n\_high\_olfactory}) reveals these features to capture conventionalized vulgar expressions like \textit{trash}, \textit{shit}, or \textit{fuck}, rather than a literal smell or sound relation of the text.
On the pragmatic/discourse level, we find surviving features related to complexity and specificity (\texttt{avg\_synsets\_noun}, \texttt{n\_polysyllables}), and stereotypical or derogatory descriptions of people such as \textit{their culture} or \textit{the Black Guy}, indicated by a relatively high number of determiners (\texttt{n\_det}).

We find \textbf{a strong intersectional effect of ideology and age},
Fig. \ref{fig:sachdeva-interaction-ideology-age}: it shows that
towards the extremes of the ideology, the difference between age groups in terms of offensiveness ratings becomes more stark, with flipped effects (the more conservative, older people rate lower on average, and the opposite for liberals). This is relevant from a label variation, a harmful language, and a content moderation perspective, as it shows that even when individual sociodemographic proxies do not indicate systematic differences, their interaction may, pointing to the impact of the lived experiences of certain subgroups.

We also find \textbf{three interactions between linguistic features and annotator characteristics}. This, again, reveals task-specific differences: the association between ratings and lexical cues, such as emotion intensity for surprise, is dependent on ideology (\texttt{ideology.C:n\_high\_surprise}
; tokens with high surprise intensity include examples like \textit{shockingly}, or \textit{disaster}), and the association between ratings and number of hedges (tokens indicating speaker uncertainty) on education level  (\texttt{education.Q:n\_hedges}
).
Finally, we find an interaction of Hindu annotators and the number of tokens not assignable to a standard POS tag (\texttt{religionhindu:n\_x}). Inspection of items and annotations points to an artifact of data and annotator sampling: Coincidentally, items with a high number of tokens with the \textit{X} tag (non-standard words/slang, misspellings, non-Latin characters) had more Hindu annotators who tendentially rated these items as not hateful. From a modeling perspective, such an effect can be viewed as a concrete example of data quality issues that such an analysis can reveal. It is reasonable to assume that language models may pick up on such spurious signals, and thus, one may want to be aware of them in their data.


\subsection{Discussion}
While we only find one common effect, the effect of negative sentiment tokens, this is expected given different tasks, annotation guidelines, items, and annotators. Interestingly, however, both models show effects given the presence of offensive, hateful, and vulgar words, validating the informativeness of such lexical signals found by \citet{rizzi-etal-2025-bunch} for related tasks.
Overall, the findings point to dataset, annotator set, and task-specific tendencies, urging caution when applying findings of one specific configuration to another.
Finally, the surviving interaction effects of annotator characteristics and linguistic features underline the importance of taking the variation between annotators and items, and their interdependences into account.


\section{Simulating New Annotators}
\label{sec:ex2-new-annotators}
Given the large number of annotators per item, \textsc{D3CODE} is well-suited to assess whether tendencies in annotation behavior persist when we collect more annotations on the same items from annotators with similar distributions of annotator characteristics to the original annotator population.
We randomly sample half of the annotators. Our halved samples retain similar distributions of annotator characteristics and numbers of annotations per item. Per half, we fit one model.

\subsection{Results}
\begin{figure}
    \centering
    \includegraphics[width=\linewidth]{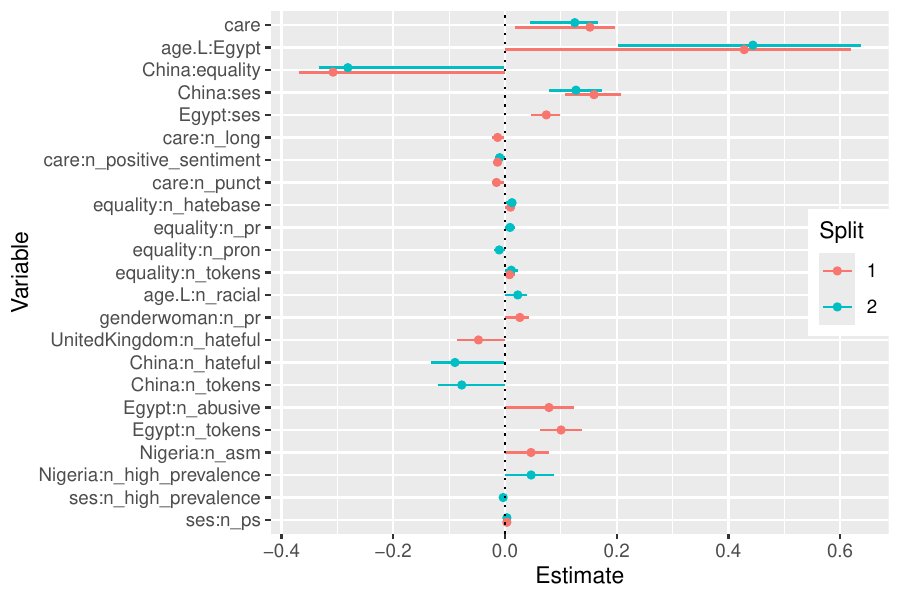}
    \caption{Posterior estimates for the surviving effects for the two halves of the annotators of \textsc{D3CODE}.}
    \label{fig:davani-surv}
\end{figure}

Figure \ref{fig:davani-surv} shows the posterior estimates for the surviving effects\footnote{Note that the effect sizes may not be strictly comparable, since moral foundations such as \texttt{care} are not standardized, while linguistic features are.} for both sets of annotators for \textsc{D3CODE}.
By itself, \textbf{one annotator characteristic survives in both models}, the moral foundation dimension \textit{care}. The moral foundations questionnaire differs from other annotator characteristics insofar as it surveys moral intuitions from people rather than measuring innate characteristics such as age or gender. Care, specifically, scores high for annotators who consider protecting vulnerable individuals from emotional and physical harm as very important, and is associated with empathy and compassion \citep{graham-2013-moral}.
For both models, we find \textbf{no surviving linguistic features}. 

We find \textbf{four intersectional effects}: For annotators from Egypt
with age and self-reported socio-economic status (SES),
and for annotators from China with SES and the moral foundation of equality, which scores high for annotators who believe that all people should be treated equally.

Across both models, \textbf{roughly 40\% of the surviving effects are interactions between lexical cue features and annotator characteristics}. We take this as a strong reason to take complex relationships of personal identity and lived experiences, and the perceptions of certain texts into account. 
For example, women
may rate items with slurs related to prostitution higher because they are gendered and target women, as \texttt{genderwoman:n\_pr} hints at.
This makes a case for considering label variation as a signal rather than noise: If women differ systematically for certain items, this should not get flattened away by label aggregation. 
Similarly, from a content moderation/modeling perspective, such perceptions of targeted groups may be the focus of interest \citep{fleisig-etal-2023-majority}.

Interactions of the annotator characteristics, moral foundations, SES, or country, with several linguistic features related to length and complexity (e.g., \texttt{care:n\_long\_words} and \texttt{Egypt:n\_tokens})
point to a complex relationship of items and annotators. On the one hand, from a label variation and harmful language research perspective, they may point to specific items that are particularly prone to varied perceptions, for instance, longer texts. On the other hand, if such interactions are only present for certain samples of annotators, but not necessarily for the general population (e.g., \texttt{Egypt:n\_abusive} is only a survivor for the first half of the annotators), this may be problematic from a modeling and content moderation perspective: A model working well on a "representative" sample of the target annotator population may still not be well-equiped for the patterns of another sample, or the population overall. This raises questions about how to model within-group variation, and which learned patterns to consider generalizable. 

\subsection{Discussion}
The results show that roughly half of the surviving effects can be found for both models. This indicates that, given the same text items, similar distributions of annotator characteristics, and the same guidelines, we can expect some level of aligned labeling behavior. The other half of the surviving effects, however, points to a certain level of intra-group variation. Overall, our results indicate that label variation, to a large extent, may be driven by individual differences for certain items than by annotator characteristics or certain items alone. This is relevant from multiple angles: From a label variation perspective, this underlines the necessity to look beyond individual proxies and understand annotation as an interactive process between annotators and items to disentangle variation. From a modeling and content moderation perspective, this raises the question of who labeled the items we are training on, and of their linguistic tendencies.

    

\section{Simulating New Annotators and Items}
\label{sec:ex3-new-batches}

\begin{figure}
    \centering
    \includegraphics[width=\linewidth]{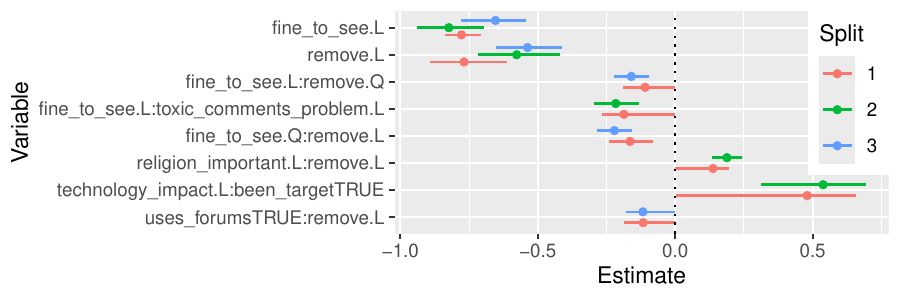}
    \caption{Posterior estimates for surviving effects of \textsc{CTDP}. We show effects that survive for $\geq2$ subsets.}
    \label{fig:kumar-surv}
\end{figure}

Given the high number of items and annotators, \textsc{CTDP} is fitting to simulate a batched annotation scenario to assess whether annotation behavior tendencies hold across batches, i.e., on previously unannotated items with completely new annotators.

We reconstruct the batches by dividing the data into subsets of items that share the same $k$ annotator IDs.
We divide \textsc{CTDP} into subsets of 300 each. We run one model on each of 3 randomly drawn subsets, totaling $\sim16\%$ of the full dataset.

\subsection{Results}
Figure \ref{fig:kumar-surv} shows effects that survive for at least two splits. Full results are presented in \S \ref{sec:ex3-full-res}.  
The only \textbf{two common surviving annotator characteristics} across all three splits are annotator attitudes towards specific items. The finer an annotator is to see a given item, the lower the toxicity annotation (\texttt{fine\_to\_see.L}). Conversely, the more an annotator thinks a post should be removed, the higher the toxicity annotation (\texttt{remove.L})\footnote{The effects are negative here because the scale is flipped; "This comment should be removed" is the lowest level, while "This comment should be allowed" is the highest.}. These annotator characteristics differ from sociodemographic proxies or general attitudes and moral foundations insofar as they measure item-specific perceptions and attitudes. Conceptually, they are thus similar to item feature-annotator characteristics interactions. 

We find \textbf{no linguistic features among the surviving effects} in $\geq2$ splits.
All \textbf{six of the surviving interactions }include two annotator characteristics. 
Two of the interactions reveal differences between groups along axes of general attitudes with item-specific attitudes (\texttt{toxic\_comments\_problem}, whether annotators think that toxic problems online are a problem interacts with \texttt{fine\_to\_see}, and \texttt{religion\_importance} interacts with \texttt{remove}). Furthermore, two reflect differences in lived realities and experiences, as shown by the interaction of whether an annotator has been the target of toxic comments and to what extent they think technology has a positive impact on society (\texttt{technology\_impact.L:been\_targetTrue}),
and whether they personally use online forums and to what extent they think a given item should be removed (\texttt{uses\_forumsTRUE:remove.L}).

\subsection{Discussion}
While there are no linguistic features or interactions with them among the surviving effects common to at least two models, the two main surviving effects
conceptually share some overlap. In contrast to sociodemographic variables like gender or general attitudes, they fundamentally reflect item-specific attitudes of annotators. 
Overall, the survivors point to the utility of phenomenon- and item-specific attributes reflecting the lived experiences and interactive perceptions in the annotation. Such attributes, as the results of this analysis in comparison with the two previous analyses indicate, may account for more variance in annotation behavior than broad sociodemographic features or even linguistic-sociodemographic interactions.
This setup particularly lends itself to content moderation, as the phenomenon and item-specific annotator attitudes allow for capturing nuanced perceptions of when and why a given person may rate an item as toxic.

\section{Conclusion}
In this work, we presented a series of analyses on four unaggregated harmful language datasets. Using multilevel Bayesian models with a rich set of linguistic features, annotator characteristics, and their interactions, we found that, while there were differences across datasets, interactions of item features and annotator characteristics (Sections \ref{sec:ex1-comparison} \& \ref{sec:ex2-new-annotators}) or item-specific attitude effects (Section \ref{sec:ex3-new-batches}) were present across analyses. This has consequences for data collection efforts:
Firstly, some items may need more annotations by a more diverse or focused set of annotators, while others may be reasonably uncontroversial across annotator groups. For example, for items containing slurs targeted at women, we may want to make sure to get them annotated by many women of different backgrounds. Secondly, our results point to the need for reflection on which annotator characteristics and attitudes should be collected, particularly on an item level. This depends on the purpose a given dataset is collected for, and which questions it is supposed to answer.
Finally, since data defines all parts of the NLP system lifecycle, our findings urge for engaging critically with the assumptions of what annotated data aims to capture and their links to model behavior, both in training and evaluation scenarios.

\clearpage

\section*{Limitations}
While our work assesses annotation variation in a principled manner, taking into account the structure of annotations and predictors on two levels, and their interactions, our work is limited in multiple regards.

Firstly, we only assess one set of tasks, offensive/hateful language detection, on one language, English. While our findings do not claim any universality, we still urge caution, given that findings may not transfer across languages, tasks, and domains.

Secondly, our assessment is limited to available annotator and item characteristics. There may be arbitrarily more, some of which may not be practically and reliably measurable (e.g., annotator mood, or short distractions). In a similar vein, our work does not account for batch-level effects\footnote{We note that this limitation directly arises from how data is documented: None of the datasets we assess includes batch IDs, and they may not always be easily reconstructable.}, intra-annotator agreement \citep{abercrombie-etal-2023-temporal}, the impact of the annotation setup (e.g., the annotation environment \citep{kern-etal-2023-annotation}, the order in which items are shown \citep{beck-etal-2024-order}), or the effect of the annotation itself. For instance, verbally thinking about one's assessment may impact the annotator's own emotional/intuitive response, given that language may modulate perception and cognition \citep{lupyan2012linguistically}.
Different formats may lead to different response distributions \citep[c.f. findings from survey methodology showing systematic differences between numbers and labelings of response options, ][]{WEIJTERS2010236}.

Moreover, annotator selection practices may impact the interpretation of findings. While over a whole dataset, annotator socio-demographics may be reflective of the whole target population (e.g., residents of a given country, or English speakers), this is not the case for any given batch and item\footnote{We note that, likely, this is due to costs, as having a set of annotators with socio-demographics reflective of the whole target population necessarily requires a high number of annotators per batch/item.}. This may have a considerable impact on the estimated effects, as many batches and items will only ever be annotated by the socio-demographic majority groups. Especially in tasks like hate speech detection, this comes with real implications for groups targeted by hateful and offensive language \citep[c.f. ][]{fleisig-etal-2023-majority}.

The extent of exploration is limited by our available resources and, more fundamentally, by available implementations\footnote{We discuss observations on these limitations and how we handle them in Appendix \ref{sec:observations}.}. In our experiments, we hit implementation limits in how complex a model can be. 
Given that, thus, we are reducing the sample size for some datasets by splitting them (Experiment 3, Section \ref{sec:ex3-new-batches}), it is very likely that the posteriors are broader than when using the full datasets. As such, it is possible that for the smaller-sample scenarios, some effects do not survive that would have survived in the full-data scenario.
Including random slopes and not only random intercepts may be informative and can be argued for, in our setup and given our resources, but it is infeasible. Similarly, item features never occur in isolation, and interactions of uncorrelated features may reveal interesting items. Due to    
Comparisons between coefficients in Experiment 2 (Section \ref{sec:ex2-new-annotators}) are not meaningful since they are not all standardized (continuous annotator characteristics are not standardized). The interpretations, however, would be limited even if they were, given that an effect of \textit{one standard deviation from the mean of the moral foundation care}, for example, is hard to interpret, and even harder to compare to effects of other such variables. Given that this is an ongoing discussion, whether and when coefficients are strictly comparable, we urge caution when trying to compare coefficients and deliberately refrain from it in this work. 

Finally, we note that our work is only one of many ways to analyze factors of variation. And at that, it is quite a conservative approach, especially with respect to what we consider a surviving effect, and how we operationalize the exploration.  
Decisions at all steps may impact findings. We discuss decision rationales and alternatives further in Appendix \ref{sec:alternatives}.

\section*{Ethical Considerations}

The work presented in this paper belongs to the perspectivist framework, whose agenda we support fully. We are, however, aware that the claim that multiple (and many) annotations per item should be collected can be problematic for research groups with limited funding -- and this should not become an economic bottleneck. We believe, however, that works like ours show that statistical analysis of existing datasets (without collecting new data) can provide fundamental insights. Moreover, our methodology, bringing together the \textit{who} and the \textit{what}, can also inform a more efficient way of using annotation budget -- one that exploits systematic patterns of interaction between annotator and item characteristics to improve dataset quality by focusing on specific sets of items.

\section*{Contributions}
In the following, we list the contributions of each author of this paper according to the CRediT taxonomy\footnote{\url{https://credit.niso.org/}}.

\begin{itemize}
    \item \textbf{Conceptualization:}\\Maximilian Maurer, Maximilian Linde, Gabriella Lapesa
    \item \textbf{Data Curation:}\\Maximilian Maurer, Maximilian Linde
    \item \textbf{Formal Analysis:}\\Maximilian Maurer, Maximilian Linde
    \item \textbf{Funding Acquisition:}\\All authors are employed in household positions at GESIS and used associated funds.
    \item \textbf{Investigation:}\\Maximilian Maurer, Maximilian Linde, Gabriella Lapesa
    \item \textbf{Methodology:}\\Maximilian Maurer, Maximilian Linde, Gabriella Lapesa
    \item \textbf{Project Administration:}\\Maximilian Maurer, Maximilian Maurer, Gabriella Lapesa
    \item \textbf{Resources:}\\All authors are employed in household positions at GESIS and used associated resources.
    \item \textbf{Software:}\\Maximilian Maurer, Maximilian Linde
    \item \textbf{Supervision:}\\Gabriella Lapesa
    \item \textbf{Validation:}\\Maximilian Maurer, Maximilian Linde, Gabriella Lapesa
    \item \textbf{Visualization:}\\Maximilian Maurer
    \item \textbf{Writing:}\\Maximilian Maurer, Maximilian Linde, Gabriella Lapesa
\end{itemize}

\bibliography{latex/custom}

\appendix
\input{latex/sections/appendices/ling-features-full}
\input{latex/sections/appendices/lexicons}
\input{latex/sections/appendices/preprocessing-full}
\input{latex/sections/appendices/full-results-ex3}
\input{latex/sections/appendices/examples}

\input{latex/sections/appendices/preselection}
\input{latex/sections/appendices/glossary}

\input{latex/sections/appendices/alternatives}
\input{latex/sections/appendices/implementation}
\input{latex/sections/appendices/popquorn-reproduction}
\input{latex/sections/appendices/comp-resources}
\input{latex/sections/appendices/alternative_modeling}

\input{latex/sections/appendices/interactions}

\FloatBarrier
\end{document}

%% file: latex/sections/introduction.tex

\begin{figure}
    \centering
    \includegraphics[width=\linewidth]{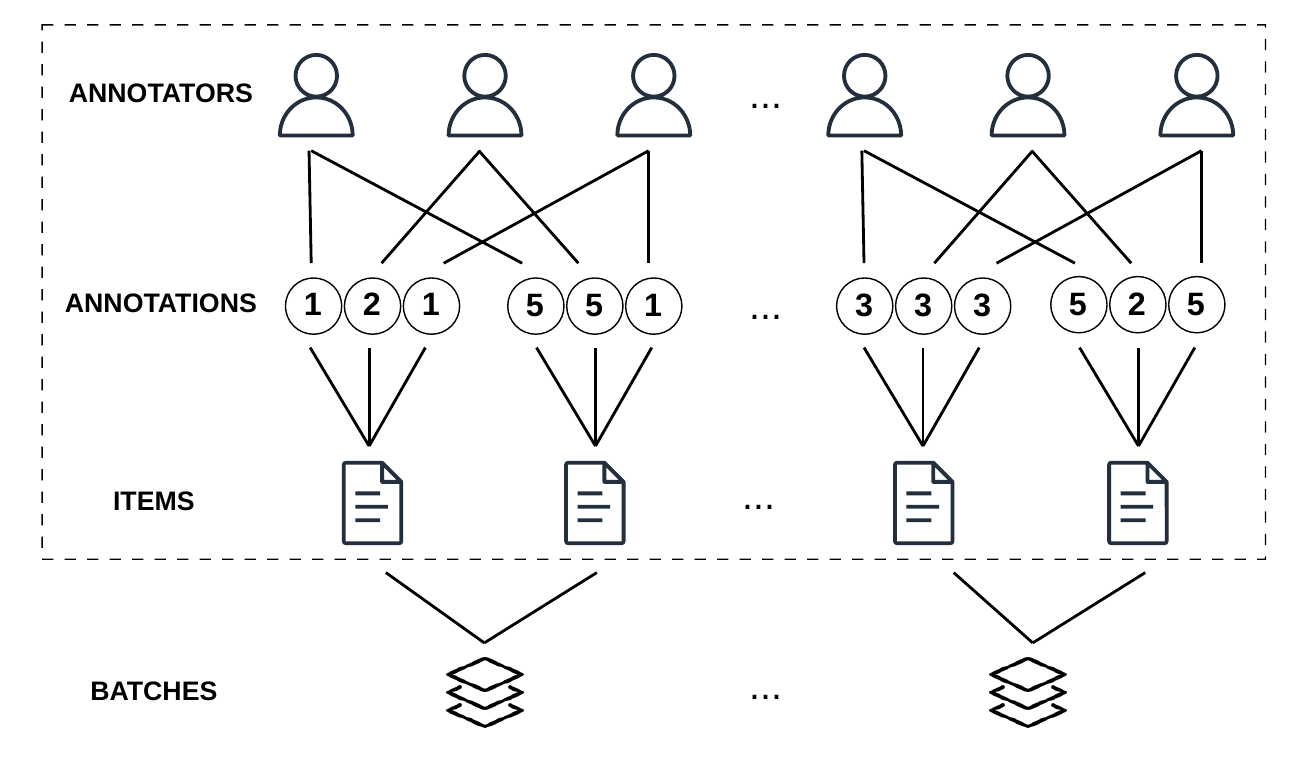}
    \caption{
    Cross-classified data structure for ordinal text annotations: Each annotation belongs to one unique annotator/item combination. Each item is part of a batch. We focus on the structure within the dashed box.}
    \label{fig:structure-annotations}
\end{figure}

In recent years, calls for considering annotation disagreement \citep{basile-etal-2021-need} and embracing annotation variation \citep[\textit{data perspectivism},][]{cabitza2023toward} have led to the introduction of \textit{disaggregated} corpora including individual annotators' labeling decisions, and modeling approaches accounting for this variation beyond aggregating them into a single gold label \citep[][among others]{davani-etal-2022-dealing, weerasooriya-etal-2023-disagreement}. In tasks where the annotator characteristics are of particular interest, \textit{perspectivist} works reveal a mixed picture: 
While some works find that including annotator sociodemographics improves the modeling of annotation variation \citep{kocon2021offensive, wan2023everyone, tahaei-bergler-2024-analysis}, others do not find convincing evidence that it does \citep{orlikowski-etal-2023-ecological, sun-etal-2025-sociodemographic}.

We argue that this discrepancy and the design of existing works reveal gaps in addressing the main underlying question: \textbf{Who} differs in their perception of \textbf{what} \citep[cf.][]{sap-etal-2022-annotators}?

Firstly, the assessment of this question is impacted by the structure of how annotations are usually conducted:
As exemplified in Figure \ref{fig:structure-annotations}, annotations follow a cross-classified structure, meaning that annotations are simultaneously grouped by items and annotators (two non-nested factors)
: each item is annotated by multiple annotators, and each annotator annotates multiple items.
One can reasonably assume systematic variation on both levels.  Most works assessing annotation variation, however, do not account for this, resulting in limited generalizability and comparability of the findings.

Secondly, while analysis approaches such as the ones mentioned above provide a starting point for the community to address the who, the what remains largely ignored. This gap is particularly noteworthy given that annotation is an interactive process between text items of varying linguistic composition and annotators, whose identities are more complex than individual sociodemographic proxies such as gender \citep[c.f.][]{orlikowski-etal-2023-ecological}. This may lead to variations between annotator groups, but only for certain text items. 


\textbf{Given that it is unclear whether annotator- and item-level characteristics are predictive of annotation behavior, we propose to analyze disaggregated data in a principled way to reveal factors of interest when dealing with annotation variation in subjective tasks.} We argue that this is a vital step before training or testing any language model-based system, as it provides important pointers to the information we might want to include in such a system
and what model behavior to expect.

In this work, we thus conduct the largest annotation variation analysis to date, spanning \textbf{four disaggregated harmful communication datasets}, 
containing a total of \textbf{>25k items}, \textbf{>8k annotators}, and \textbf{>205k annotations}.
We take both annotator characteristics (up to 19) and item features (over 300) into account, as well as interactions among annotator features and between the annotator and item features. 
We use Bayesian multilevel regression models to find the most impactful and relevant out of up to 
5,264 fixed effects.
We account for the partially cross-classified structure by including random intercepts for annotators and items.

On the annotator side, we take available sociodemographic (e.g., age) and attitudes (e.g., whether annotators think hate speech is a problem)
and two-way interactions between them into account. This allows for assessing intersectional effects.


On the item side, we look at domain-specific lexical signals, as well as a broad set of general characteristics, ranging from morphosyntactical to psycholinguistic features. To assess who differs in their perception of what, we include interactions between the annotator and item features.

We conduct exploratory analyses
in three realistic scenarios\footnote{While these scenarios are realistic, they were partially motivated by restrictions due to computational resources and limits of existing implementations (see App. \ref{sec:used-resources} \& \ref{sec:observations}).}:
(i) Comparing effects in related tasks with different conceptualizations, annotation guidelines, items, and annotators. This aims at finding potentially more general effects in related phenomena (Section \ref{sec:ex1-comparison}). (ii) Comparing demographically similar annotator groups: two annotator groups with very similar distributions of annotator characteristics, annotating the same items with the same conceptualization and annotation guidelines. This can be viewed as a simulation of collecting more annotators for items for which one already has annotations
(Section \ref{sec:ex2-new-annotators}). (iii) Comparing different sets of batches. This can be viewed as datasets collected with different annotators for different items 
but using the same conceptualization and annotation guidelines (Section \ref{sec:ex3-new-batches}).

Our contributions are two-fold:
(1) On a methodological level, we conduct a principled in-depth analysis of large disaggregated datasets. We discuss relevant questions, assumptions, and decisions at each step of our analyses. In doing so, we hope to contribute to establishing best practices in the field when analyzing such datasets.
(2) On a substantial level, we provide, to the best of our knowledge, the first assessment of annotation behavior from a linguistic, annotator-item interaction, and intersectional perspective for harmful language datasets.

Answering to the question "who annotates what" is relevant for multiple NLP research communities: the harmful language detection community benefits from the analysis of these reference datasets and may find more insights in the effect patterns we discovered. The human label variation community may benefit from considering both the who and the what, for disentangling variation, but also targeted (re-)annotation.
Finally, the modeling/content moderation community benefits from insights informing about differences of tendencies for specific item-annotator combinations, and for identifying potential spurious confounders.

%% file: latex/sections/related_work.tex

Besides annotation errors, taxonomies \citep{basile2021a, uma2021, zhang2023} have identified three high-level sources of annotation variation: reasons stemming from the annotator, the items, and the annotation guidelines and settings.


On the annotators' side, subjectivity and individual differences in sociodemographic backgrounds and attitudes
have received particular attention. While some works find significant impacts of the country of residence \citep{lee-etal-2024-exploring-cross}, race \citep{larimore-etal-2021-reconsidering}, gender, and age \cite{pei-jurgens-2023-annotator}, others do not find such differences \citep{biester-etal-2022-analyzing, sap-etal-2022-annotators}. Modeling approaches including sociodemographic information reflect this mixed picture \citep{wan2023everyone,  orlikowski-etal-2023-ecological, tahaei-bergler-2024-analysis, beck-etal-2024-sensitivity, orlikowski-etal-2025-beyond, sun-etal-2025-sociodemographic}.
\citet{homan-etal-2024-intersectionality} argue that identities are more complex than individual demographic characteristics, so they investigate intersectional effects, and find differences between intersectional groups, particularly for \textit{race} and \textit{gender}.

In contrast to the potential reasons for annotation variation on the annotators' side, work on reasons on the item side remains largely theoretical. A noteworthy exception is \citet{rizzi-etal-2025-bunch}, who find that for hate speech, certain lexical items are indicative of disagreement in harmful language. Theoretical taxonomies \citep{uma2021, basile2021a} name item difficulty, (missing) context, and linguistic ambiguities on all levels, as well as uncommon words, code switching, and neologisms as potential reasons for annotation variation. They, however no not conduct extensive analyses or considering interactions.
For online toxicity, \citet{zhang2023} add domain-specific reasons such as obfuscated racism. 
Moreover, they hint at reasons due to an interaction between annotator- and item-level factors, such as varying sensitivity to
lexical signals. \citet{sap-etal-2022-annotators} draw further attention to factors relating to both annotators and items. They analyze the impact of demographics and attitudes on the annotator side, and dialect differences on the item side. They find that more conservative annotators rate anti-Black statements lower in toxicity compared to other annotator groups, but African American English items higher. Similarly, \citet{larimore-etal-2021-reconsidering} find interactions between annotator \textit{race} and racially charged lexical signals such as the \textit{n-word}, or charged topics such as police brutality. 

\begin{table}
    \centering
    \resizebox{\linewidth}{!}{
    \begin{tabular}{llcccc}
    \toprule
         
         \multirow{2}{*}{Work} & \multirow{2}{*}{Analysis Method} & \multicolumn{2}{c}{Features} & \multicolumn{2}{c}{Random Intercepts}\\
         \cmidrule(lr){3-4}\cmidrule(lr){5-6}
         && Annotators & Items & Annotators & Items\\
         \midrule
         \citet{wan2023everyone} & LM-based Classification & \cmark & \xmark & \multicolumn{2}{c}{\textbf{N/A}}\\
         \citet{orlikowski-etal-2023-ecological} & LM-based Classification & \cmark & \xmark & \multicolumn{2}{c}{\textbf{N/A}}\\
         \citet{rizzi-etal-2025-bunch} & LM-based Classification & \xmark & \cmark & \multicolumn{2}{c}{\textbf{N/A}}\\
         \citet{mostafazadeh-davani-etal-2024-d3code} & Descriptive Analysis & \cmark & \xmark & \multicolumn{2}{c}{\textbf{N/A}}\\
         \citet{kumar2021diversity} & Logistic Regression & \cmark & \xmark & \xmark & \xmark \\
         \citet{larimore-etal-2021-reconsidering} & Linear Regression & \cmark & \cmark & \xmark & \xmark\\
         \citet{pei-jurgens-2023-annotator} & Multilevel Regression & \cmark & \xmark & \xmark & \cmark \\ 
         \textbf{Ours} & Multilevel Regression & \checkmark & \checkmark & \checkmark & \checkmark \\
         \bottomrule
    \end{tabular}}
    \caption{Comparative overview of the analyses done in related work vs. ours: do they consider annotator characteristics and interpretable item features? For regression-based experiments, do they account for variation of annotators and items with random intercepts?}
    \label{tab:related-overview}
\end{table}

\paragraph{Research Gap} Drawing on these findings,
in this work, we view annotation as an interactive process between annotators and items. As such, we do not only include item-level and annotator-level features but also consider interactions between the two levels. In contrast to related studies, we take the partially cross-classified data structure into account.
On the \textit{what}-level, we include both characteristics that can be assumed to be indicative of the phenomenon (e.g., lexical signals idiosyncratic to the phenomenon) and broader phenomenon-independent features (e.g., lexical richness or uncertainty markers). The latter may point to more general linguistic indicators of annotation variation.  Table \ref{tab:related-overview} provides an overview of the key analysis differences between prior work and our work.

%% file: latex/sections/data.tex
We use four English harmful language (hate speech, toxic and offensive language) datasets fulfilling the desiderata of having (a) unaggregated annotations (i.e., individual annotators' labeling decisions are available), (b) annotator characteristics (socio-demographic attributes, attitudes), and (c) fine-grained annotations (3-/5-point Likert-scales).



\noindent 
\textbf{CTDP.} The corpus for toxic content classification for a diversity of perspectives collected by \citet{kumar2021diversity} consists of social media comments from Reddit, Twitter, and 4chan annotated for offensiveness on a 5-point scale. It has the annotator characteristics of gender, age, ethnicity, politics, religious importance, LGBTQ+, and parent status.


\noindent
\textbf{MHS.} The Measuring Hate Speech Corpus \citep{sachdeva-etal-2022-measuring} consists of social media comments annotated for hate speech on a 3-point scale. It provides the annotator characteristics race, gender, sexuality, religion, education, and income, and a social media and hatespeech questionnaire asking about general and item-specific attitudes.


\noindent
\textbf{POPQUORN.} The POPQUORN corpus \citep{pei-jurgens-2023-annotator} contains annotations for the tasks of question-answering, offensiveness rating, text rewriting/style transfer, and politeness rating. We use the offensiveness rating portion, consisting of Reddit comments labeled on a 5-point scale for offensiveness. It provides the annotator characteristics of gender, age, ethnicity, politics, occupation, and education



\noindent
\textbf{D3CODE.} The D3CODE dataset \citep{mostafazadeh-davani-etal-2024-d3code} contains online comments annotated for toxicity on a 5-point scale. It provides the annotator characteristics of gender, age, country of residence, geo-cultural region, moral foundations, and perceived socio-economic status.

%% file: latex/sections/linguistic_features.tex
To assess item-side factors of annotation variation, we use general linguistic features and domain-specific lexical signals.
We extract linguistic features using \texttt{elfen} \citep{maurer-2025-elfen}. Overall, we consider 327 features from the 11 provided feature areas. 
We provide a full overview of the used features in Appendix \ref{sec:linguistic-features-full}. 



To find domain-specific lexical signals such as vulgarity or hateful slurs, we use the English portions of the Harassment Corpus \citep{harassment-corpus}, which contains harassment-related words in six categories (\textit{sexual, appearance-related, intellectual, political, racial}, and \textit{combined}), Hurtlex \citep{bassignana2018hurtlex}, which contains \textit{words to hurt} in three course-grained categories (\textit{Negative stereotypes}, \textit{hate words and slurs beyond stereotypes}, and \textit{Other words and insults}) and 17 fine-grained categories,
\citet{wiegand-etal-2018-inducing}'s base lexicon of abusive words, and Hatebase\footnote{\href{https://hatebase.org/}{https://hatebase.org/}}.
We extract the number of offensive words from the four hate speech lexicons and the number of words from fine-grained categories per item. 
We list all domain-specific lexicon-based features in Appendix \ref{sec:lexicons}.

%% file: latex/sections/appendices/ling-features-full.tex
\section{Full Overview of Linguistic Features}
\label{sec:linguistic-features-full}
In the following, we describe the linguistic features used in this work. For a full overview, check \citet{maurer-2025-elfen}.

\paragraph{Surface-Level Features} We extract the sequence length (characters; both with and without whitespaces, \texttt{raw\_sequence\_length}, and \texttt{n\_characters}), number of tokens (\texttt{n\_tokens}), sentences (\texttt{n\_sentences}), types (\texttt{n\_types}), lemmas (\texttt{n\_lemmas}), long words (over six characters, \texttt{n\_long\_words}), the number of tokens per sentence (\texttt{tokens\_per\_sentence}), characters per sentence (\texttt{characters\_per\_sentence}), and average word length (\texttt{avg\_word\_length}).

\paragraph{Readability Features} We extract the Gunning fog index (\texttt{gunning\_fog}), ARI (\texttt{ari}), Flesch reading ease (\texttt{flesch\_reading\_ease}), and Flesch-Kincaid grade level \citep[\texttt{flesch\_kincaid\_grade,}][]{kincaid1975derivation}, the Cole-Liau index \citep[\texttt{cli,}][]{coleman1975computer}, SMOG \citep[\texttt{smog},][]{mc1969smog}, LIX \citep[\texttt{lix},][]{björnsson1968läsbarhet}, and RIX \citep[\texttt{rix},][]{anderson1981analysing}, the number of syllables in an item (\texttt{n\_syllables}), words with only one syllable (\texttt{n\_monosyllables}), and words with more than two syllables (\texttt{n\_polysyllables}).

\paragraph{Psycholinguistic Norm Features} We extract the average rating across the tokens of an item (\texttt{avg\_\{norm\}}), the average standard deviation in the human ratings across the tokens of an item (\texttt{avg\_std\_\{norm\}}), the number of tokens in an item with a high rating (upper third of the ordinal scale, \texttt{\texttt{n\_high\_\{norm\}}}), the number of tokens in an item with a low rating (lower third of the ordinal scale, \texttt{\texttt{n\_low\_\{norm\}}}), and the number of tokens with a particularly high standard deviation (spanning over multiple thirds of the scale, \texttt{\texttt{n\_high\_std\_\{norm\}}})
We use concreteness \citep{brysbaert2014concreteness}, marked by \texttt{concreteness} (e.g., in \texttt{avg\_concreteness}), word prevalence \citep{brysbaert2019word},
marked by \texttt{prevalence}, Age-of-Acquisition norms \citep{kuperman2012age}, marked by \texttt{aoa}, Socialness norms \citep{diveica2023quantifying}, marked by \texttt{socialness}, Iconicity norms \citep{winter2024iconicity}, marked by \texttt{iconicity}, and Sensorimotor norms \citep{lynott2020lancaster}, per perceptual modalities (e.g. visual) and action effectors (e.g. arm/hand), marked by \texttt{\{modality|effector\}} (e.g., \texttt{avg\_arm}).

\paragraph{Part-of-Speech Features.} We extract the number of tokens per POS tag (e.g., the number of nouns, \texttt{n\_noun}), the number of lexical tokens (nouns, verbs, adjectives, and adverbs, \texttt{n\_lexical\_tokens}), and the POS variability (number of different POS tags relative to the number of tokens, \texttt{pos\_variability}).

\paragraph{Lexical Richness Measures} We extract of the type-token ratio (\texttt{ttr}) \citep{templin1957certain}, root TTR \citep[\texttt{rttr},][]{guiraud1954caractères}, corrected TTR\citep[\texttt{cttr}, ][]{carroll1964language}, Herdan's C \citep[\texttt{herdan\_c, }][]{herdan1964quantitative}, Summer's TTR (\texttt{summer\_index, }), Dugast's Uber index\citep[\texttt{dugast\_u, }][]{dugast1978}, Maas' TTR \citep[\texttt{maas\_index, }][]{mass1972zusammenhang}, Yule's $K$ \citep[\texttt{yule\_k}, ][]{yule1944statistical}, Herdan's $V_m$ \citep[\texttt{herdan\_v}][]{herdan1955new}, Simpson's $D$ \citep[\texttt{simpsons\_d, }][]{simpson1949measurement}, mean segmental TTR \citep[\texttt{msttr, }][]{richards1997quantifying}, moving average TTR \citep[\texttt{mattr, }][]{covington2010cutting}, measure of textual lexical diversity  
 \citep[\texttt{mtld, }][]{mccarthy2010mtld}, and the hypergeometric distribution diversity \citep[\texttt{hdd, }][]{mccarthy2007vocd, mccarthy2010mtld}, the local and global numbers of hapax (dis)legomena (\texttt{n\_hapax\_legomena}, \texttt{n\_global\_token\_hapax\_legomena}), Sichel's S \citep[\texttt{sichel\_s}, ][]{sichel1975distribution}, and the lexical density (\texttt{lexical\_density}).

\paragraph{Morphological Features.} We extract the number of tokens with a given morphological feature for all available universal dependencies morpho-syntactic features \citep{de-marneffe-etal-2021-universal}, marked in the format \texttt{n\_\{pos\}\_\{attribute\}\_\{feature\}} (e.g., the number of singular nouns, \texttt{n\_NOUN\_Number\_Sing}).

\paragraph{Information-Theoretic Features.} We extract the compressibility (\texttt{compressibility}) and Shannon entropy per item (\texttt{entropy}).

\paragraph{Dependency Features} we extract the number of dependency relation types \citep[according to Universal Dependencies,][]{de-marneffe-etal-2021-universal}, marked in the format \texttt{n\_dependency\_\{type\}} (e.g., \texttt{n\_dependency\_nsubj}), the number of noun chunks in the text (\texttt{noun\_chunks}), the tree width (\texttt{tree\_width}), the tree depth (\texttt{tree\_depth}), the tree branching factor (\texttt{branching\_factor}), and the ramification factor (\texttt{ramification\_factor}).

\paragraph{Semantic Features.} We extract the average size of the synsets (\texttt{avg\_synsets}), the number of tokens with a large synset (more than four senses; \texttt{n\_high\_synsets}), and the number of tokens with a small synset (less than three senses, \texttt{n\_low\_synsets}) for nouns, adjectives, and verbs, respectively, and overall.
We extract the number of hedges (\texttt{n\_hedges}), i.e., expressions that indicate speakers uncertainty, for example "probably", "maybe", "i think", etc..

\paragraph{Named Entity Features.} We extract the number of named entities overall (\texttt{n\_entities}) and per entity type (e.g., \texttt{n\_fac}, i.e., facilities like buildings, airports and the like).

\paragraph{Emotion and Sentiment Features.} We use the NRC-VAD lexicon \citep{mohammad-2018-obtaining} for valence, arousal, and dominance, the NRC emotion intensity lexicon \citep{mohammad-2018-word} for the emotion intensity per basic emotion (anger, anticipation, disgust, fear, joy, sadness, surprise, trust), and
the NRC word-emotion association lexicon \citep{mohammad-turney-2010-emotions, mohammad2013crowdsourcing} for sentiment. Per emotion dimension, we extract the average rating per item (\texttt{avg\_\{emotion\}}, \texttt{avg\_\{valence|arousal|dominance\}}), the number of tokens with a high rating (\texttt{n\_high\_\{emotion\}}, \texttt{n\_high\_\{valence|arousal|dominance\}}), and the number of tokens with a low rating (\texttt{n\_low\_\{emotion\}}, \texttt{n\_low\_\{valence|arousal|dominance\}}). For sentiment, per item, we extract the number of positive and negative sentiments (\texttt{n\_\{positive|negative\}\_sentiment}, and the difference between them, normalized by the total number of tokens in the item (\texttt{sentiment\_score}).

%% file: latex/sections/appendices/lexicons.tex
\section{Domain-specific Lexicons}
\label{sec:lexicons}

\begin{table*}[!t]
    \centering
    \resizebox{\linewidth}{!}{
    \begin{tabular}{lll}
         \toprule
         Source & Feature & Explanation \\
         \midrule
         Hatebase & n\_hatebase & Number of tokens found on Hatebase\\
         \hline
         Abusive Words & n\_abusive & Number of tokens found in the Abusive Words lexicon\\
         \hline
         Hurtlex & n\_ps & Number of negative stereotype/ethnic slur tokens\\
         & n\_rci & Number of location/demonym tokens\\
         & n\_pa & Number of profession/occupation tokens\\
         & n\_ddf & Number of tokens related to physical disabilities and diversity\\
         & n\_ddp & Number of tokens related to cognitive disabilities and diversity\\
         & n\_dmc & Number of tokens related to moral and behavioral defects\\
         & n\_rci & Number of tokens related to physical disabilities and diversity\\
         & n\_is & Number of tokens related to social and economic disadvantage\\
         & n\_or & Number of tokens related to plants\\
         & n\_an & Number of tokens related to animals\\
         & n\_asm & Number of tokens related to male genitalia\\
         & n\_asf & Number of tokens related to female genitalia\\
         & n\_ps & Number of tokens related to prostitution\\
         & n\_om & Number of tokens related to homosexuality\\
         & n\_qas & Number of tokens with potentially negative connotations\\
         & n\_cds & Number of derogatory tokens\\
         & n\_re & Number of tokens related to felonies, crime, and immoral behavior\\
         & n\_svp & Number of tokens related to the seven deadly sins of the Christian tradition\\
         \hline
         Harassment Lexicon & n\_generic & Number of tokens related to harassment\\
         & n\_sexual & Number of tokens related to sexual harassment\\
         & n\_appearance & Number of tokens related to appearance-related harassment\\
         & n\_racial & Number of tokens related to racial harassment\\
         & n\_intelligence & Number of tokens related to intellectual harassment\\
         & n\_politics & Number of tokens related to political harassment\\\midrule
        & n\_hateful & Number of tokens in the union of all lexicons\\
         \bottomrule
    \end{tabular}}
    \caption{Domain-specific lexicon-based features with explanations for them.}
    \label{tab:lexicons}
\end{table*}

Table \ref{tab:lexicons} lists the features per source, including a short explanation.

%% file: latex/sections/appendices/preprocessing-full.tex
\section{Preprocessing: Full Details}
\label{sec:preprocessing-full}
In the following, we describe the preprocessing of the linguistic features and annotator characteristics used in this work's analyses.

\paragraph{Linguistic Features}
We preprocess the linguistic features in two steps:
(1) If the feature is an occurrence-count feature (e.g., number of hedges), we normalize by the number of tokens in the respective item (i.e., we divide the respective occurrence count by the number of tokens in the item). This allows us to analyze relative trends rather than raw frequencies. For instance, longer texts can generally be expected to have more hedges, but the interesting items are those with a high number of hedges relative to their length. 
(2) We standardize all linguistic features to have a mean of $0$ and a standard deviation of $1$.

\paragraph{Annotator Characteristics}
Before pre-processing annotator characteristics, we remove missing answers and "prefer not to answer" (or similar) responses. We remove all annotators with conflicting characteristics for the same annotator ID. This ensures the expected data structure, i.e., each annotator ID corresponds to exactly one annotator. While this potentially removes cases of reasonable changes over the span of the respective annotation period (e.g., an annotator turning 35, moving into another age group), this ensures removing annotators that are almost certainly spammers (e.g., annotators changing all the characteristics).

We harmonize the socio-demographic variables \textit{gender} (by mapping to \textit{male}, \textit{female}, and \textit{diverse}), and \textit{education} by mapping the respective datasets' scheme to the international standard classification of education levels \citep[ISCED 2011, ][]{isced2011}. We harmonize \textsc{MHS} with \textsc{POPQUORN} by coding raw age in years into the respective age ranges (e.g. $26\to25\text{-}29$).

\texttt{CTDP} and \texttt{MHS} allow for arbitrary \textit{race}, and \texttt{CTDP} for arbitrary \textit{sexuality} and \textit{religion} answers, leading to self-described identities like \textit{Buddhist, Christian and Atheist} for the same person. While this points to people wanting to express their more complex identities, it leads to an explosion of the number of categories, with most of them being rather sparse. For \textit{race}, we thus only keep categories that involve exactly one \textit{race} (e.g. \textit{White} or \textit{Asian}) and the five most frequent categories involving two. The more complex categories are mapped to a catch-all \textit{multiracial} category.
For religion and sexuality, we only keep annotators belonging to categories involving exactly one religion and sexuality, respectively. In consequence, we drop $\sim 2\%$ of annotators for \texttt{MHS} assigning multi-religious categories and multiple sexualities to themselves\footnote{We observe co-occurrences especially for \textit{bisexual} and \textit{straight} or \textit{gay}, potentially indicating these annotators would've wished for more fine-grained preference options.}.

In \texttt{D3CODE} we drop the \textit{cultural region} variable and only keep the more fine-grained geo-cultural indicator \textit{country}.

We keep all other annotator characteristics and code them accordingly. Table \ref{tab:sd_overview} in Appendix \ref{sec:annotator-characteristics-details} provides an overview of all annotator characteristics used in the following analyses with their variable type (e.g., nominal or ordinal) and the chosen reference level for nominal variables.

\section{Details Annotator Characteristics}
\label{sec:annotator-characteristics-details}

\begin{table}[H]
    \centering
    \resizebox{\linewidth}{!}{
    \begin{tabular}{llll}
        \toprule
         Feature & Type & Reference & Dataset \\
         \midrule
         Gender & nominal & \textit{male} & all\\
         Age& ordinal &  & all\\
         Education & ordinal &  & \texttt{CTDP}, \texttt{POPQUORN}, \texttt{MHS}\\
         Race & nominal & \textit{white} & \texttt{CTDP}, \texttt{POPQUORN}, \texttt{MHS}\\
         Political ideology & ordinal &  & \texttt{MHS}\\
         Political affiliation & nominal & \textit{Liberal} & \texttt{CTDP}\\
         Socio-economic status & ordinal &  & \texttt{D3CODE}, \texttt{POPQUORN}\\ 
         Moral foundations & interval &  & \texttt{D3CODE}\\
         Country & nominal & \textit{USA} & \texttt{D3CODE}\\
         Media usage & nominal & \textit{no} & \texttt{CTDP}\\
         Task-specific questionnaire & ordinal & & \texttt{CTDP}\\
         & nominal & \textit{no} & \texttt{CTDP}\\
         Occupation & nominal & \textit{employed full-time} & \texttt{POPQUORN}\\
         LGBTQ status & nominal & \textit{heterosexual} & \texttt{CTDP}\\
         Trans status & nominal & \textit{no} & \texttt{MHS}\\
         \bottomrule
    \end{tabular}}
    \caption{Annotator characteristics per dataset with data type and reference level for nominal variables.
    }
    \label{tab:sd_overview}
\end{table}

Table \ref{tab:sd_overview} shows annotator characteristics per dataset with data type and reference level for nominal variables.

%% file: latex/sections/appendices/full-results-ex3.tex
\section{Full Results for CTDP}
\label{sec:ex3-full-res}
\begin{figure}
    \centering
    \includegraphics[width=\linewidth]{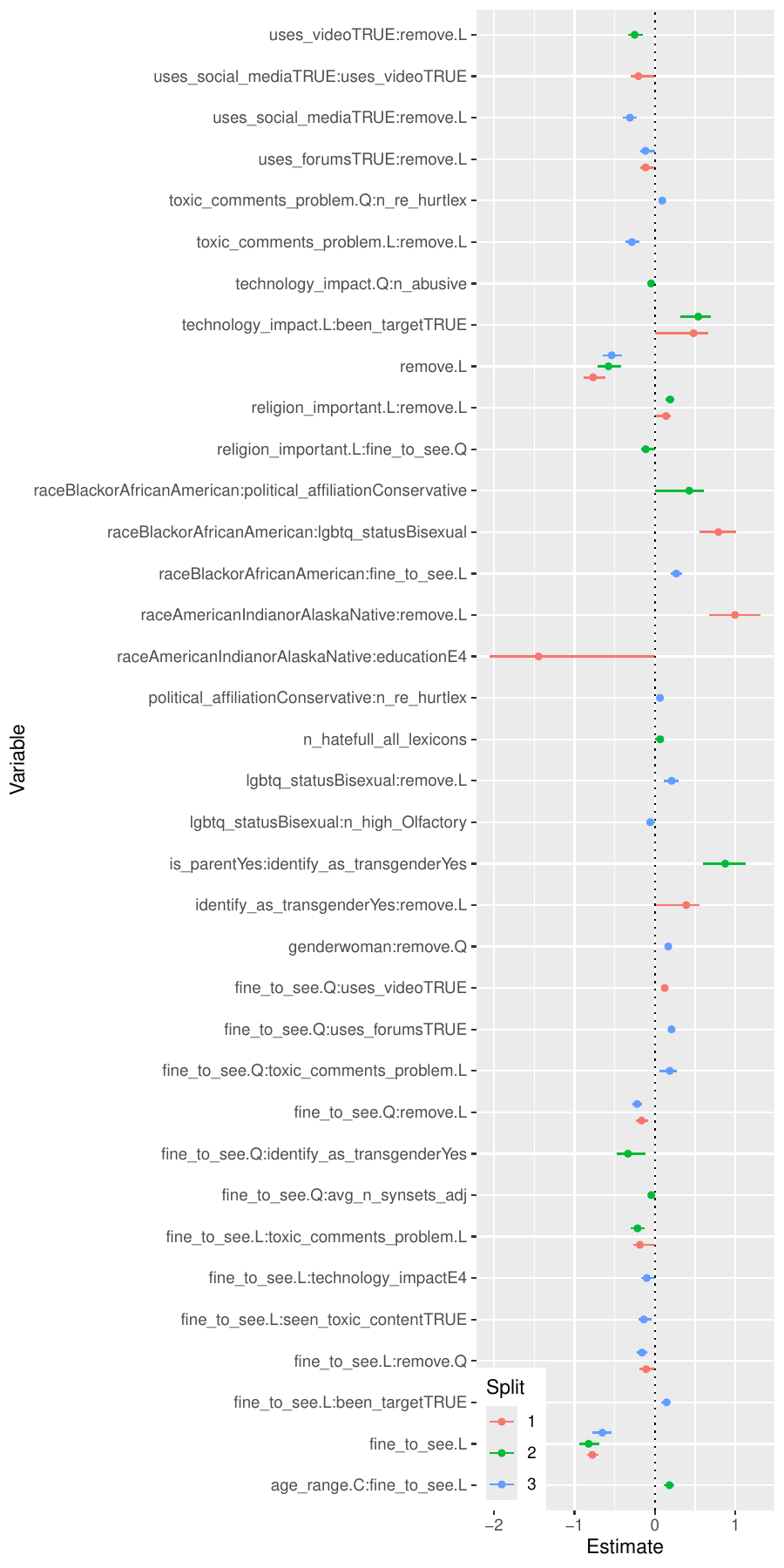}
    \caption{Posterior estimates for surviving effects of CTDP. }
    \label{fig:kumar-surv-all}
\end{figure}

Figure \ref{fig:kumar-surv-all} shows all survivors across the analysis in Section \ref{sec:ex3-new-batches}.

%% file: latex/sections/appendices/examples.tex
\section{Text Examples}
\label{sec:text-examples}
Figure \ref{fig:example-1} contains examples for items from POPQUORN containing a relatively high number of words related to moral/behavioral deficiencies (\texttt{n\_dmc}), Figure \ref{fig:example-1} examples for items from POPQUORN containing a relatively high number of named entities of type nationalities, religious or political groups(\texttt{n\_norp}). 
Figure \ref{fig:example-3} contains examples for items from MHS containing a relatively high number of tokens with a high olfactory grounding (\texttt{n\_high\_Olfactory}), Figure \ref{fig:example-4} contains examples for items from MHS containing a relatively high number of tokens with a high auditory grounding (\texttt{n\_high\_Auditory}). 

\begin{figure}
    \hfill
    \centering
    \begin{subfigure}{0.9\columnwidth}
            \small
    Ok. Thanks for your opinion.. I don’t know if the story is true or not.. if true.. they should be prosecuted! Don’t enter my home illegally! No excuse for \colorbox{cyan}{evil} on part of authorities. ... I’m not naive to believe LAR’s are ALL angelic beings. 78.2\% of Washington Post info is biased, unverified and \colorbox{cyan}{misleading}. There’s your debate.
    \normalfont{}
    \subcaption{}
    \end{subfigure}
    \\
    \vspace{0.5cm}
    \hfill
    \begin{subfigure}{0.9\columnwidth}
        \small
    I don't think you can argue this very well, because someone who has less \colorbox{cyan}{morals} or \colorbox{cyan}{ethics} than another person is not excluded from acting \colorbox{cyan}{morally} or ethicallt, but someone who does have morals and ethics would not act \colorbox{cyan}{immoral} or \colorbox{cyan}{unethical}. For this reason the person who has less morals or ethics has a larger set of options when making money compared to the more moral/ethical, so they will always make equal or better than the moral person. I think a more difficult question is whether acting immorally makes you more money than acting morally.
    \normalfont{}
    \subcaption{}
    \end{subfigure}
    \\
    \vspace{0.5cm}
    \hfill
    \begin{subfigure}{0.9\columnwidth}
        \small
    Well, they are likely \colorbox{cyan}{filthy} rich...
    \normalfont{}
    \subcaption{}
    \end{subfigure}
    \caption{Example items from POPQUORN containing a relatively high number of words related to moral/behavioral deficiencies (\texttt{n\_dmc} colored in \colorbox{cyan}{cyan}).}
    \label{fig:example-1}
\end{figure}

\begin{figure}
    \hfill
    \centering
    \begin{subfigure}{0.9\columnwidth}
            \small
    Don't worry. \colorbox{cyan}{Israel} has already told the \colorbox{cyan}{UN} there will be no investigation. Gotta love that \colorbox{cyan}{jew} privilege.
    \normalfont{}
    \subcaption{}
    \end{subfigure}
    \\
    \vspace{0.5cm}
    \hfill
    \begin{subfigure}{0.9\columnwidth}
        \small
    Congrats on the 1:30 \colorbox{cyan}{Israeli}/\colorbox{cyan}{Palestinian} casualty ratio. Hamas must be patting themselves on their backs and looking for a repeat of that success.
    \normalfont{}
    \subcaption{}
    \end{subfigure}
    \\
    \vspace{0.5cm}
    \hfill
    \begin{subfigure}{0.9\columnwidth}
        \small
    A less exhaustive background check would risk accepting \colorbox{cyan}{Muslims} on the flight, which would completely defeat the whole point of a \colorbox{cyan}{Muslim}-free flight.
    \normalfont{}
    \subcaption{}
    \end{subfigure}
    \\
    \vspace{0.5cm}
    \hfill
    \begin{subfigure}{0.9\columnwidth}
        \small
    I guess I can say the same about \colorbox{cyan}{Muslims}? Just waiting for y'all to learn that they do in \colorbox{cyan}{Muslim} countries, it makes the \colorbox{cyan}{Catholic} Church look like Disneyland
    \normalfont{}
    \subcaption{}
    \end{subfigure}
    \caption{Example items from POPQUORN containing a relatively high number of named entities of type nationalities, religious or political groups(\texttt{n\_norp}, colored in \colorbox{cyan}{cyan}).}
    \label{fig:example-2}
\end{figure}

\begin{figure}
    \hfill
    \centering
    \begin{subfigure}{0.9\columnwidth}
            \small
    Hahaha I love \colorbox{cyan}{trash} bag nazi Americans. Have fun getting shot you inbred piece of \colorbox{cyan}{shit}
    \normalfont{}
    \subcaption{}
    \end{subfigure}
    \\
    \vspace{0.5cm}
    \hfill
    \begin{subfigure}{0.9\columnwidth}
        \small
    Some real \colorbox{cyan}{nasty} women over here, the \colorbox{cyan}{nastiest}. Sad!
    \normalfont{}
    \subcaption{}
    \end{subfigure}
    \\
    \vspace{0.5cm}
    \hfill
    \begin{subfigure}{0.9\columnwidth}
        \small
    talk \colorbox{cyan}{shit} to my face so i can rock your \colorbox{cyan}{shit} pussy
    \normalfont{}
    \subcaption{}
    \end{subfigure}
    \caption{Example items from MHS containing a relatively high number of tokens with a high olfactory grounding (\texttt{n\_high\_Olfactory}, colored in \colorbox{cyan}{cyan}).}
    \label{fig:example-3}
\end{figure}

\begin{figure}
    \hfill
    \centering
    \begin{subfigure}{0.9\columnwidth}
            \small
    It's easy, randomly \colorbox{cyan}{click} until you lose.
    \normalfont{}
    \subcaption{}
    \end{subfigure}
    \\
    \vspace{0.5cm}
    \hfill
    \begin{subfigure}{0.9\columnwidth}
        \small
    We all know she'll try to *\colorbox{cyan}{talk} to the manager* while down in hell.
    \normalfont{}
    \subcaption{}
    \end{subfigure}
    \caption{Example items from MHS containing a relatively high number of tokens with a high auditory grounding (\texttt{n\_high\_Auditory}, colored in \colorbox{cyan}{cyan}).}
    \label{fig:example-4}
\end{figure}

%% file: latex/sections/appendices/preselection.tex
\section{Linguistic Feature Preselection: Example}
\label{sec:clustering}
To showcase our selection procedure as described in Section \ref{sec:methods}, we go through a full selection procedure in the following for \texttt{POPQUORN}.

We start with 327 features, which we first filter by pairwise Pearson correlation, arriving at 106 features, each of which is correlated lower than $0.5$ with any of the other features.
The remaining 225 features are clustered to select features of interest in the correlation clusters.

\begin{figure}[H]
    \texttt{['tokens\_per\_sentence', 'tree\_width', 'tree\_depth', 'tree\_branching', 'entropy', 'mtld', 'yule\_k', 'herdan\_v', 'n\_polysyllables', 'flesch\_reading\_ease', 'flesch\_kincaid\_grade', 'gunning\_fog', 'ari', 'smog', 'lix', 'rix']}
    \caption{Example: Cluster 6 in the linguistic feature preselection procedure for \texttt{POPQUORN}.}
    \label{fig:cluster-6-popquorn}
\end{figure}

For instance, one of the clusters is shown in Figure \ref{fig:cluster-6-popquorn}. The features in this cluster are all related to readability and syntactic complexity. In this case, we pick the Flesch reading ease, as it is a well-used readability measure.

\begin{figure}[H]
    \texttt{['avg\_word\_length', 'n\_long\_words', 'n\_NOUN\_Number\_Sing', 'n\_PRON\_PronType\_Prs', 'n\_PRON\_Number\_Sing', 'n\_PRON\_Case\_Nom', 'n\_PRON\_Person\_1', 'n\_ADJ\_Degree\_Pos', 'n\_PROPN\_Number\_Sing', 'n\_dependency\_amod', 'n\_dependency\_compound', 'n\_dependency\_nsubj', 'n\_lexical\_tokens', 'lexical\_density', 'n\_adj', 'n\_noun', 'n\_pron', 'n\_propn', 'avg\_aoa', 'n\_high\_aoa', 'avg\_sd\_aoa', 'cli', 'avg\_n\_synsets', 'avg\_n\_synsets\_verb', 'n\_low\_synsets', 'n\_high\_synsets', 'n\_entities', 'n\_person', 'n\_entities\_token\_ratio', 'n\_entities\_sentence\_ratio', 'n\_global\_token\_hapax\_legomena', 'n\_global\_token\_hapax\_dislegomena', 'global\_sichel\_s', 'n\_global\_lemma\_hapax\_legomena', 'n\_global\_lemma\_hapax\_dislegomena']}
    \caption{Example: Cluster 2 in the linguistic feature preselection procedure for \texttt{POPQUORN}.}
    \label{fig:cluster-2-popquorn}
\end{figure}

Naturally, as theoretical equivalence is only one reason for high correlations, not all of the clusters will contain clearly-cut, theoretically equivalent features. For instance, in the cluster shown in Figure \ref{fig:cluster-2-popquorn}. We find reasonably related but not necessarily theoretically equivalent features. For example, the number of nouns (\texttt{n\_noun}), the number of named entities (\texttt{n\_entities}), the number of adjectives (\texttt{n\_adj}), and the number of pronouns (\texttt{n\_pron}) clearly have systematic relationships. In cases like this, we pick a feature allowing for an intuitive interpretation. In this concrete case, we pick the number of nouns. 

In the end, we arrive at 113 linguistic features we use in the regression modeling of \texttt{POPQUORN}.

%% file: latex/sections/appendices/glossary.tex
\section{Term Glossary}
\label{sec:glossary}

The aim of this appendix section is to clarify the meaning of key terms employed in the paper, providing the reader with  conceptual coordinates that are necessary for a (deeper) understanding of the statistical concepts employed in our narrative, and yet too extensive to find room in the main paper.

\paragraph{(Partially) cross-classified data structure} Datasets regularly follow certain structures. For examples, pupils are nested within classes, which are in turn nested within schools, which are in turn nested within neighborhoods, which are in turn nested within cities. These nested structures evoke dependencies, which should be taken into account. That is, it is likely that pupils within one class are more similar compared to other pupils. In other situations, observations can be ascribed to a combination of factors. More specifically, a rating is made for one specific item by one specific annotator. Here, annotator ID and item ID are factorially crossed. When all combinations of annotator ID and item ID exist, we call this a fully cross-classified data structure. In contrast, when only part of all combinations of annotator ID and item ID exist, we call this a partially cross-classified data structure. In our analyses, we encounter partially cross-classified data structures because not all annotators rated all items or, vice versa, not all items were rated by all annotators.

\paragraph{Frequentist vs. Bayesian (regression)} Regression analysis is a statistical method for investigating relationships among variables. It can be used for purposes of prediction and/or explanation. In this paper, annotation behavior (i.e, ratings of hatefulness on a scale from 1 to 5) is predicted on the basis of several linguistic features of the texts and annotator characteristics.

Frequentist and Bayesian form two separate philosophical frameworks to statistics. One key difference is the notion of probability. While frequentists see probability as a long-run frequency of events (e.g., how often do I get heads if I flip a coin 100,000 times?), Bayesians see probability as a degree of belief or plausibility (e.g., I believe that there is a 90\% change that it will rain tomorrow). As such, the interpretation of results differs. The treatment of probability as a degree of belief opens the possibility to incorporate prior information/beliefs into a regression model. In that sense, Bayesian statistics updates our prior beliefs by incorporating the data that we have into a posterior belief (i.e., what we should belief after seeing the data).


Whether to pick a frequentist or Bayesian analysis is often a matter of philosophical/statistical preference. However, some situations particularly lend themselves to a Bayesian framework. For example, it can be argued that data with few observations profit from more stable estimates in a Bayesian framework because of the incorporation of prior knowledge. As another example, if the data is high-dimensional, with more predictors than observations, and the goal is to find the most important predictors, then a Bayesian regression with a horseshoe prior works well.
This is precisely the use case of our analyses. While some of the datasets we employ have a large number of observations (at least for some datasets), our large number of predictors (annotator attributes, linguistic features, interactions) creates a situation in which the amount of observations is limited compared to the number of parameters. In this scenario, adopting a Bayesian approach allows for a more stable estimation of the surviving effects, thanks to the strong impact of uncertainty in the regularization process. In practical terms, given the observation/parameter ratio, this approach allows us to be more confident in the effects we observe, than we would have been in a frequentist framework. An additional practical reason is the fact that the Bayesian framework has progressed a lot in the computational optimization for hierarchical models (compensating for the higher computational cost).


\paragraph{Outcome (predicted, dependent) vs. predictor (independent)} In regression analysis, we do not only investigate simple relationships among variables. Instead, we try to predict an outcome variable (also called dependent variable) based on one or more predictor variables (also called independent variables). For example, in this paper we attempt to predict annotation behavior on the basis of several linguistic features of the texts as well as annotator characteristics.

\paragraph{Main effect vs. interaction effect} In a regression model, there are different types of effects that might be of interest. Main effects relate to the effect of a variable on the outcome, keeping other independent variables constant at some value. For instance, what is the effect of annotator age on annotation behavior, keeping all the other independent variables constant at some value? 
If we employ annotator age or presence of hateful words as main effects in the prediction of annotation behavior, we will learn to what extent an increase in the age of the annotator or in the number of hateful words in an item  corresponds to changes in offensiveness ratings (keeping all the other independent variables constant at some value). 
Take for example the plot in Figure \ref{fig:popquorn-surv}, containing the estimates for the main effects (and interactions) for the POPQUORN dataset. Positive  estimates indicate that, when a predictor has a higher value (if continuous, e.g., number of hateful words), the model has identified a tendency for annotations to be higher (i.e., higher degree of offensiveness). Negative estimates indicate the opposite. 

In contrast, interaction effects demonstrate whether the effect of one independent variable on the dependent variable depends on the value of another independent variable, holding all other independent variables constant at $0$.
We standardized the predictors, so that the mean value is at $0$. As such, the interpretation of interaction effects is meaningful because keeping all other independent variables constant at $0$ translates to keeping all other independent variables constant at the mean. 
For example, is the effect of age on annotation behavior different depending on the presence of hateful words in the annotated items? To fully understand interaction effects, it is not enough to inspect Figure \ref{fig:popquorn-surv}: in this case, the estimate is telling us that the impact of the number of hateful words on offensiveness ratings depends on age sub-groups of our annotators. This is precisely the effect displayed in Figure     \ref{fig:pei-interaction-age-hatefull_all}. The plot shows how the effect on annotation behavior between three chosen values for the presence of hateful words is different for the age groups. Recall that the variables are standardized, so $0$ means that an item is in the range of the average number of hurtful words across all items in the dataset, $1$ that it is one standard deviation above that value, and $-1$ that it is one standard deviation below. The interaction shows that as annotator age increases, the effect of the number of hateful words on annotation behavior becomes more pronounced.

\paragraph{Fixed vs. random effect} Many types of data follow a hierarchical structure. For example, in our scenario, annotations belong to one specific combination of annotator ID and item ID (see glossary `(Partially) cross-classified data structure'). This creates dependencies/correlations within clusters/groups that should be taken into account in most cases when creating statistical models. More specifically, we would expect that annotators have different baseline levels of annotating items; one annotator might be very sensitive to hateful items while another annotator might not be. The same applies to items. These natural tendencies can be modeled in a regression by introducing random intercepts (i.e., one type of random effect). The resulting parameter estimate shows the natural tendency of annotators or items to vary at baseline.

In practical terms, this means that when estimating the fixed effects for our predictors, we assume that each individual annotator has a different starting point when annotating hatefulness/offensiveness: this is the baseline 
from which they start off, technically in the model, the random intercept. The stronger the variance between the "baselines" of each annotator in the dataset, the stronger the impact of the corresponding random effect. Given the subjectivity of the tasks at hand, it is crucial to "factor out" this type of individual variation when making generalizations about the data being annotated. More generally, we consider it good practice to consider random effects in any task, even the less subjective ones, given that we have a multilevel data structure.
This is true even if the random effects turn out to be weak (i.e., our annotators have a similar baseline in annotation behavior).

The other type of random effect would be random slopes, which would show whether there are different relationships between predictors and the outcome, depending on annotator ID or item ID. We did not include random slopes in our models. 

In contrast to random effects, the familiar fixed effects (e.g., the relationship between age and annotation behavior) constitute systematic population-average effects.


\paragraph{Prior} The use of priors is a distinctive feature of Bayesian statistics as opposed to frequentist statistics. Priors are probability distributions that, in most cases, are supposed to mirror previous knowledge (i.e., before seeing the data) about some assertion, such as a parameter (e.g., the probability that a coin lands heads).
A simple example is the toss of a coin: Before tossing a specific coin, we already know that the coin is probably fair and brings up heads and tails in equal proportions. As such, we would place more certainty on 50\%/50\% compared to 10\%/90\%. This knowledge (and our certainty thereof) can be incorporated into the prior. Bayesian modeling can then be conceived as updating our prior knowledge/suspicion by incorporating what the data tells us into a posterior knowledge (i.e., what we should believe after having incorporated what we learned from the data into our prior knowledge). To continue with the coin toss example, assume that prior to tossing the coin you have no reason to doubt its fairness. Now you toss the coin $100$ times and heads is observed $99$ times. Based on this we would be well-advised to rethink and update our beliefs about the fairness of the coin. How much we should adjust our beliefs depends on how strong our beliefs were initially. If our beliefs were quite vague, we should update more towards biasedness; if our beliefs were strong, we should update less.

Even though this toy example provides an intuition about Bayesian statistics, it is worthwhile to consider an example that is more scientifically relevant and where the prior is informed by previous study results. Imagine that previous studies have shown that there is a positive effect of age on annotator ratings. This alone gives us information that the prior should be focused on positive effects. We could further look at the effect estimates and their uncertainties to refine our prior. That is, we could more precisely set more emphasis on the particular range of effects that was found in previous studies. Thought this way, the prior summarizes previous scientific knowledge and our study and our data serve to update this knowledge.

There are certain situations, however, where priors are strategically used for a certain purpose. Here, the prior does not reflect prior beliefs anymore. In our analyses, we use the Horseshoe prior for strong regularization. That means that weak and uncertain effects get pushed toward $0$; only stronger and more certain effects remain mostly untouched.



%% file: latex/sections/appendices/alternatives.tex
\section{Methodological Choices \& Alternatives}
\label{sec:alternatives}
At given points of our analysis, we decided on one of multiple alternatives. For full transparency, we discuss reasoning and alternatives in the following subsections.

\subsection{Preprocessing}

\paragraph{Removing Missing/Non-Answers}
Given that one focus of our analyses is to explore whether there are systematic effects of annotator characteristics and interactions of them with item-level features, we chose to remove missing answers and "Prefer not to answer" responses, since they make modeling more complex, especially for otherwise ordinal predictors. Furthermore, reasons why people do not want to disclose certain information about themselves may be multifaceted or complex to disentangle. 

It can, however, be argued that these missing or non-answers provide valuable information. But dealing with missing data is notoriously difficult because it is rarely clear whether data are missing completely at random, missing at random, or missing not at random. Working with missing data would involve investigating the source of missingness.

\paragraph{Using Catch-all Categories}
For very rare multi-answer categories in the annotator characteristics of \textit{race} and \textit{religion}, we chose to re-code them to catch-all categories (see Appendix \ref{sec:preprocessing-full} for more details). While such categories may reflect complex identities of annotators that will be flattened when re-coded to catch-all categories, other approaches, such as treating each answer option as its own predictor and encoding them in a one-hot manner, would increase the complexity of our models. For works particularly interested in these specific characteristics, and, for instance, in multiracial identities and interactions of individual racial backgrounds, this may be a viable alternative.

\paragraph{Harmonizing Features}
For our first analysis (Section \ref{sec:ex1-comparison}), we re-code education to match the ISCED 2011 across datasets. While this is necessary for comparability across datasets that span annotators from and residing in multiple countries, a study focused on a single country may not benefit from or require that step.

\subsection{Feature Selection}
\paragraph{More Theoretic Works}
The present work, to a large extent, is exploratory.
The presented (linguistic) feature selection method in the main body of this paper aims to remove, in terms of the correlation structure,
redundant features while still retaining as many features as possible. A more theory-driven analysis, for example, could alleviate the necessity for a partially automated feature selection procedure by carefully pre-defining which features are of interest for the given question, and eliminating the necessity to include hundreds or thousands of predictors.

\paragraph{Heuristic Selection of Features}
The manual selection of representative features after clustering in our feature selection workflow is done mostly for interpretability reasons. In principle, the inspection could be replaced with an automatic choice per cluster. This could be either done randomly or heuristically, for example, with a ranking of which features to keep over others if in doubt. Depending on the correlation structure, this could also be applied to the whole set of features instead of clusters.

\paragraph{Dimensionality Reduction}
While approaches using dimensionality reduction may be sensible for extracting latent features that are uncorrelated/orthogonal in an $n$-dimensional feature space, one of our goals in selecting linguistic features is to retain interpretability for individual predictors. Lower-dimensional latent features may be preferable in works where computational performance and prediction are vital, or notions of distance are of interest, as they provide a more compact representation that is more straightforward to inspect and causes lower complexity in models using them.

\subsection{Modeling Decisions}
\paragraph{Random Slopes}
We did not model random slopes because there is no strong theoretical basis for this, and it would have been computationally infeasible. A more focused analysis with fewer features could explore whether features vary in their relationship to annotation behavior across groups.

\paragraph{Measurement Level of the Annotation Behavior and Linkage Functions}
In our analyses, we treat annotation behavior as a continuous variable, as we assume the underlying mental construct to be quasi-continuous. Since it is measured on a Likert scale, however, one can argue that it should not be modeled as a continuous variable. To check robustness, we also model it as an ordinal variable using a cumulative likelihood with a probit link function \citep{BxrknerVuorre2019} in pilot experiments in Appendix \ref{sec:model-robustness} for \textsc{POPQUORN}.

%% file: latex/sections/appendices/implementation.tex
\section{Implementation Details}
\label{sec:implementation-details}
\subsection{Preprocessing}
Throughout the preprocessing pipeline, we use \texttt{Python 3.10.16} and \texttt{polars}.
Our linguistic feature extraction uses \texttt{elfen 1.0.2}. Our feature filtering procedure uses \texttt{scipy 1.14.1} and \texttt{numpy 1.26.4}.

We use \texttt{tidyverse 2.0.0} in \texttt{R 4.3.3}, and \texttt{polars} in \texttt{Python 3.10.16} for preprocessing the annotator characteristics.

\subsection{Regression Modeling}
We implement the regression models in \texttt{R 4.3.3} using \texttt{brms 2.23.0} with the \texttt{cmdstanr 0.9.0} backend (\texttt{cmdstan 2.37.0}).

We use the default parameters for the horseshoe prior and the default priors in brms \citep{brms} for all other parameters.

\subsubsection{Model Formulation}
Annotation behavior,
is modeled as a function of the main effects of the linguistic and annotator features
(\verb|X_L| and \verb|X_S|, respectively)
, the interactions among the annotator features
(\verb|X_S:X_S|)
, and the interactions between linguistic and annotator features
(\verb|X_L:X_S|)
. To incorporate the partially cross-classified data structure, random intercepts for the items and annotators are included
(\verb=(1 | item)= and \verb=(1 | annotator)=, respectively):

\begin{verbatim}
y ~ X_L + X_S + X_S:X_S + X_L:X_S +
   (1 | item) + (1 | annotator)
\end{verbatim}

\subsubsection{Sampling Details}
For each
analysis, we use $4$ chains to sample from the posterior distribution. Each chain is initiated with $2,000$ warmup iterations, which are then discarded. After warmup, $7,500$ samples are drawn from the posterior, yielding $30,000$ samples for consideration.

\subsubsection{Number of Effects per Dataset Models}
Table \ref{tab:n-effects} shows the total number of effects per dataset model.

\begin{table}[H]
    \centering
    \begin{tabular}{lr}
         \toprule
         Dataset  & Number of Effects \\
         \midrule
         \textsc{POPQUORN} & 455 \\
         \textsc{MHS} & 677\\
         \textsc{D3CODE} & 816 \\
         \textsc{CTDP} & 1652 \\
         \bottomrule
    \end{tabular}
    \caption{Number of effects per dataset models, including main effects and interactions.}
    \label{tab:n-effects}
\end{table}

\subsection{Usage of AI Assistants}
In this work, we used GitHub Copilot for inline suggestions and Grammarly for grammatical corrections.

%% file: latex/sections/appendices/popquorn-reproduction.tex
\section{Reproduction of the Original POPQUORN Analysis}
\label{sec:reproduction-popqorn}

Table \ref{tab:reproduction} shows a reproduction of the effects for POPQUORN reported by \citet{pei-jurgens-2023-annotator} only with random intercepts for items (\texttt{(1 | item)}) and a comparison to the same model with random intercepts for both items and annotators (\texttt{(1 | item) + (1| annotator)}). Note that, in contrast to the analyses in the main text of this work, these are frequentist models. As the results show, given models with the same predictors, all significant effects in a model with only a random intercept for items disappear when a random intercept for annotators is added.
 
\begin{table*}
    \centering
    \resizebox{\textwidth}{!}{
    \begin{tabular}{l rrrr rrrr}
    \toprule
    & \multicolumn{3}{c}{\texttt{(1 | item)}} & \multicolumn{3}{c}{\texttt{(1 | item) + (1| annotator)}}\\
    \cmidrule(lr){2-4}\cmidrule(lr){5-7}
    & Coef. & Std.Err. & $P>|t|$ & Coef. & Std.Err. & $P>|t|$ \\
    \midrule
(Intercept)                                  & \textbf{2.100e+00}  & 5.294e-02 &  < 2e-16 &  \textbf{2.09092}  &  0.18002 &  <2e-16 \\
gender: Non-binary                           & \textbf{-2.346e-01} & 6.037e-02 & 0.000103 & -0.22899  &  0.21495 &  0.2878 \\
gender: Woman                                &          -2.121e-02 & 2.031e-02 & 0.296495 & -0.01472  &  0.07231 &  0.8389 \\  
race: Black or African American              &  \textbf{1.834e-01} & 4.472e-02 & 4.14e-05 &  0.18723  &  0.15972 &  0.2423 \\
race: Hispanic or Latino                     & \textbf{-4.061e-01} & 7.825e-02 & 2.14e-07 & -0.39655  &  0.27853 &  0.1558 \\
race: White                                  & \textbf{-1.052e-01} & 3.772e-02 & 0.005293 & -0.09897  &  0.13454 &  0.4627 \\
age: 18-24                                   & \textbf{-1.149e-01} & 4.236e-02 & 0.006686 & -0.12159  &  0.15064 &  0.4203 \\
age: 25-29                                   & \textbf{-2.995e-01} & 4.460e-02 & 1.98e-11 & -0.29698  &  0.15937 &  0.0636 \\
age: 30-34                                   & \textbf{-2.790e-01} & 4.208e-02 & 3.49e-11 & -0.27695  &  0.14995 &  0.0660 \\
age: 35-39                                   & \textbf{-2.555e-01} & 4.157e-02 & 8.20e-10 & -0.26162  &  0.14833 &  0.0790 \\
age: 40-44                                   & \textbf{-1.495e-01} & 4.341e-02 & 0.000574 & -0.14841  &  0.15472 &  0.3384 \\
age: 45-49                                   & \textbf{-2.012e-01} & 4.439e-02 & 5.88e-06 & -0.20009  &  0.15772 &  0.2058 \\
age: 50-54                                   & \textbf{-2.557e-01} & 4.677e-02 & 4.68e-08 & -0.26517  &  0.16685 &  0.1133 \\
age: 54-59                                   & \textbf{-1.135e-01} & 3.988e-02 & 0.004438 & -0.11315  &  0.14180 &  0.4257 \\
age: 60-64                                   & \textbf{ 1.956e-01} & 5.057e-02 & 0.000110 &  0.19039  &  0.17966 &  0.2903 \\
education: Graduate degree                   & \textbf{ 6.558e-02} & 2.667e-02 & 0.013961 &  0.06701  &  0.09489 &  0.4808 \\
education: High school diploma or equivalent &  1.502e-02 & 2.276e-02 & 0.509393 &  0.02367  &  0.08114 &  0.7708 \\
\bottomrule
    \end{tabular}}
    \caption{Reproduction of the original POPQUORN analysis with only random intercepts for items (\texttt{(1 | item)}) and a comparison with also including annotator intercepts (\texttt{(1 | item) + (1| annotator)}). We report the coefficients (Coef.), the standard deviation error (Std.Err.), and p-value ($P>|t|$). Significant estimates ($p<0.05$) are \textbf{bolded}.}
    \label{tab:reproduction}
\end{table*}

%% file: latex/sections/appendices/comp-resources.tex
\section{Used Resources}
\label{sec:used-resources}
For transparency and as an estimate for similar analyses, this section reports the computational resources that were needed for the analyses reported in the main body of this paper. We report (a) the required memory, and (b) compute times per dataset and model.
\subsection{Memory Needs}
Table \ref{tab:memory} shows the memory needs of our experiments. We note that these memory needs are specific to the parameter choices, our hardware, and the choice of the \texttt{brms} backend. For some observations and comparisons on this, see Appendix \ref{sec:observations}.

\begin{table}[H]
    \centering
    \resizebox{\linewidth}{!}{
    \begin{tabular}{llrr}
         \toprule
         Dataset & Split & RSS (in GB) & VSZ (in GB)\\
         \midrule
         \textsc{POPQUORN} & -- & 3.97 & 27.95 \\
         \midrule
         \textsc{MHS} & -- & 10.77 & 34.56 \\
         \midrule
         \multirow{2}{*}{\textsc{D3CODE}} & 1 & 22.65 & 46.75 \\
         \cmidrule{2-4}
         & 2 & 22.59 & 46.69\\
         \midrule
         \multirow{4}{*}{\textsc{CTDP}} & 1 & 8.63 & 25.49\\
         \cmidrule{2-4}
         & 2 & 8.54 & 25.46\\
         \cmidrule{2-4}
         & 3 & 9.21 & 26.36\\
         \bottomrule
    \end{tabular}}
    \caption{Used memory of each of the models in our analyses.}
    \label{tab:memory}
\end{table}

\subsection{Compute Times}
Table \ref{tab:execution-times} shows the runtimes of the models reported in the main body of this work.

\begin{table}[H]
    \centering
    \begin{tabular}{llr}
         \toprule
         Dataset & Split & Time (in Days) \\
         \midrule
         \textsc{POPQUORN} & -- & 4.23 \\
         \midrule
         \textsc{MHS} & -- & 10.59\\
         \midrule
         \multirow{2}{*}{\textsc{D3CODE}} & 1 & 11.12 \\
         \cmidrule{2-3}
         & 2 & 12.23\\
         \midrule
         \multirow{4}{*}{\textsc{CTDP}} & 1 & 5.74 \\
         \cmidrule{2-3}
         & 2 & 6.48\\
         \cmidrule{2-3}
         & 3 & 7.26\\
         \midrule
         \multicolumn{2}{c}{\textbf{Total}} & 57.65\\
         \bottomrule
    \end{tabular}
    \caption{Compute times of each of the final models used in this work and overall.}
    \label{tab:execution-times}
\end{table}

\section{Observations on Limitations of Implementations and Resources}
\label{sec:observations}
During our analyses, we were confronted with several limitations of implementations and our available resources. In the following, we discuss the most marked ones.

\textbf{The brms backend that is used can make a drastic difference.} Given the large number of predictors in our regression models, we had to balance runtime and memory requirements. Pilot experiments on earlier formulations of models for the \textsc{POPQUORN} dataset. When comparing the exact same model formulation with \texttt{cmdstanr}, the backend we use throughout the reported analyses in the paper, instead of \texttt{rstan}, we see a \textbf{60\% reduction in memory needs}. This, however, does not come without a drawback, as we see a \textbf{20\% increase in runtime.} 

Similarly, using \texttt{cmdstanr} may come with limitations with respect to how many predictors of which type (ordinal, nominal, etc.) are used, and how they are represented in intermediate steps internally. We ran into several $2^{31}-1$ bytes limitation errors that traced back to internal intermediate transformations to JSON strings. While this specific problem may pertain to the specific version of \texttt{cmdstanr} we are using, this points to the more general limitation that given large enough data and complex enough model formulations, \textbf{existing implementations may not be able to handle such analyses.}

%% file: latex/sections/appendices/alternative_modeling.tex
\section{Robustness Check: Testing different model formulations.}
\label{sec:model-robustness}
To test robustness, we ran pilot experiments on \texttt{POPQUORN} with a Gaussian likelihood and an identity link function, and a cumulative likelihood and a probit link function, with three horseshoe prior settings: (a) the default horseshoe prior, (b) a horseshoe prior with the global shrinkage parameter set to half of the default, and (c) a horseshoe prior with the student-t slab scale set to $10^{6}$.

\begin{table}[H]
    \centering
    \resizebox{\linewidth}{!}{
    \begin{tabular}{l|llllll}
         & \textbf{GA} & \textbf{GB} & \textbf{GC} & \textbf{PA} & \textbf{PB} & \textbf{PC}\\
         \hline
         \textbf{GA} & 1.000 & 0.999 & 0.999 & 0.913 & 0.912 & 0.915\\
         \textbf{GB} & 0.999 & 1.000 & 0.999 & 0.913 & 0.913 & 0.916\\
         \textbf{GC} & 0.999 & 0.999 & 1.000 & 0.913 & 0.913 & 0.915\\
         \textbf{PA} & 0.913 & 0.913 & 0.913 & 1.000 & 0.999 & 0.999\\
         \textbf{PB} & 0.912 & 0.913 & 0.913 & 0.999 & 1.000 & 0.999\\
         \textbf{PC} & 0.915 & 0.916 & 0.915 & 0.999 & 0.999 & 1.000\\
    \end{tabular}}
    \caption{Pairwise Pearson correlations between z-scored estimates of different model configurations. \textbf{G} refers to Gaussian likelihood with an identity link function, and \textbf{P} to the cumulative likelihood with a probit link function.
    \textbf{A} refers to the default horseshoe prior, \textbf{B} to a horseshoe prior with a halved global shrinkage parameter, and \textbf{C} to a horseshoe prior with the student-t slab scale set to
    $10^{6}$.
    }
    \label{tab:models-corr}
\end{table}

We compare the models by calculating the pairwise Pearson correlation between their z-scored estimates. Table \ref{tab:models-corr} shows the results. All of the combinations reach a Pearson correlation of over 0.91, indicating stable results across model formulations.
While there is no guarantee that this holds for our other datasets, we assume this to be the case and, given the time and resource requirements of each of the runs, do not run such stability comparisons for the other datasets.

%% file: latex/sections/appendices/interactions.tex
\section{Interactions}
\label{sec:interactions}


\begin{figure}
    \centering
    \includegraphics[width=\linewidth]{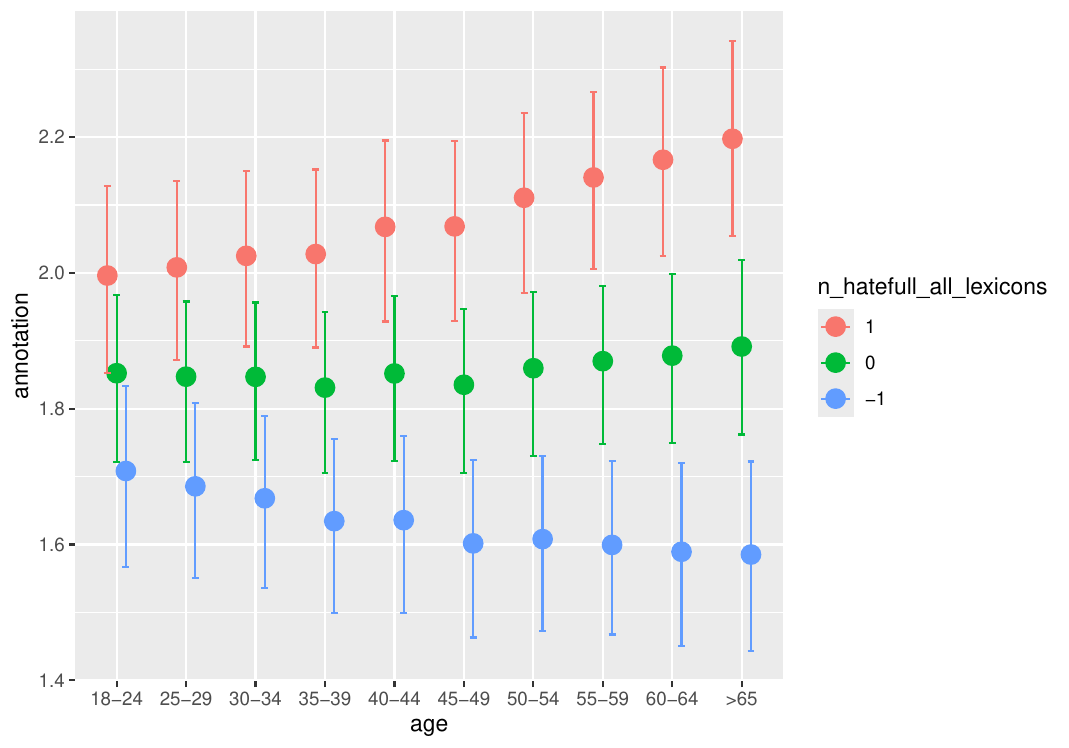}
    \caption{age:n\_hateful\_all\_lexicons (POPQUORN)}
    \label{fig:pei-interaction-age-hatefull_all}
\end{figure}



\begin{figure}
    \centering
    \includegraphics[width=\linewidth]{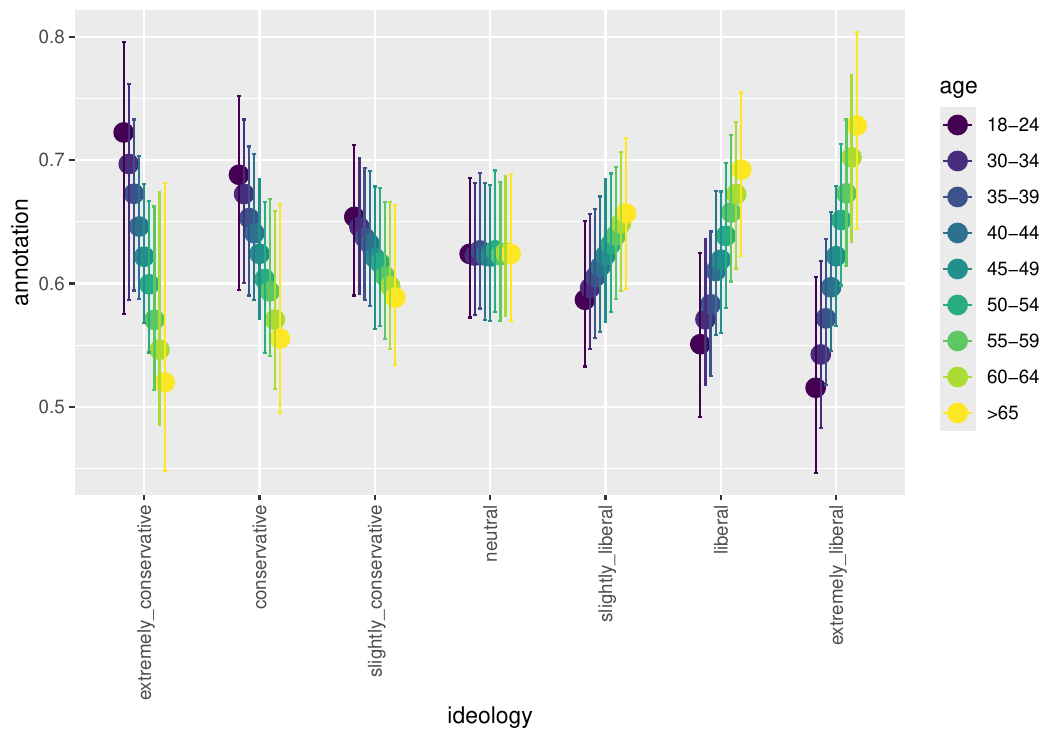}
    \caption{ideology:age (MHS)}
    \label{fig:sachdeva-interaction-ideology-age}
\end{figure}